\documentclass[10pt,twocolumn,letterpaper]{article}

\usepackage{iccv}
\usepackage{times}
\usepackage{epsfig}
\usepackage{graphicx}
\usepackage{amsmath}
\usepackage{amssymb}
\usepackage{graphicx}
\usepackage{color}
\usepackage{booktabs}
\usepackage{gensymb}
\usepackage{multirow}
\usepackage{subfig}
\usepackage{comment}
\usepackage{float}

\usepackage[ruled,vlined]{algorithm2e}
\include{pythonlisting}

\DeclareMathOperator*{\argmin}{argmin}
\DeclareMathOperator{\Pos}{Pos}
\DeclareMathOperator{\Neg}{Neg}
\DeclareMathOperator{\shp}{SH}
\DeclareMathOperator{\mhp}{MH}

\def\BState{\State\hskip-\ALG@thistlm}

\usepackage[breaklinks=true,bookmarks=false, hidelinks=true]{hyperref}

\iccvfinalcopy 

\def\iccvPaperID{1768} 

\newcommand\blfootnote[1]{%
	\begingroup
	\renewcommand\thefootnote{}\footnote{#1}%
	\addtocounter{footnote}{-1}%
	\endgroup
}

\begin{document}

\title{Explaining the Ambiguity of Object Detection and 6D Pose From Visual Data}

\author{\hspace{-3mm}Fabian Manhardt$^{1,*}$\\
\hspace{-3mm}{\tt\small fabian.manhardt@tum.de}
\and
\hspace{-3mm}Diego Martin Arroyo$^{1,*}$\\
\hspace{-3mm}{\tt\small martin.arroyo@tum.de}\\
\and
\hspace{-3mm}Christian Rupprecht$^{2}$\\
\hspace{-3mm}{\tt\small chrisr@robots.ox.ac.uk}
\and
\hspace{-3mm}Benjamin Busam$^{1,3}$\\
\hspace{-3mm}{\tt\small benjamin.busam@huawei.com}
\and
\hspace{-3mm}Tolga Birdal$^{4}$\\
\hspace{-3mm}{\tt\small tbirdal@stanford.edu}
\and
\hspace{-3mm}Nassir Navab$^{1}$\\
\hspace{-3mm}{\tt\small nassir.navab@tum.de}
\and
\hspace{-3mm}Federico Tombari$^{1,5}$\\
\hspace{-3mm}{\tt\small tombari@in.tum.de}\\
\and
$^{1}$Technical University of Munich \and \hspace{-5mm}$^{2}$University of Oxford \and \hspace{-5mm}$^{3}$Huawei \and \hspace{-5mm}$^{4}$Stanford University \and \hspace{-5mm}$^{5}$Google
}

\maketitle
\blfootnote{* The first two authors contributed equally to this work.}

\begin{abstract}
	3D object detection and pose estimation from a single image are two inherently ambiguous problems. Oftentimes, objects appear similar from different viewpoints due to shape symmetries, occlusion and repetitive textures. This ambiguity in both detection and pose estimation means that an object instance can be perfectly described by several different poses and even classes. In this work we propose to explicitly deal with this uncertainty. For each object instance we predict multiple pose and class outcomes to estimate the specific pose distribution generated by symmetries and repetitive textures. The distribution collapses to a single outcome when the visual appearance uniquely identifies just one valid pose. We show the benefits of our approach which provides not only a better explanation for pose ambiguity, but also a higher accuracy in terms of pose estimation.
\end{abstract}


\section{Introduction}

Driven by deep learning, image-based object detection has recently made a tremendous leap forward in both accuracy as well as efficiency~\cite{Ren2015,He2016,Liu2016,Redmon2016}. An emerging research direction in this field is the estimation of the object's pose in 3D space over the existing 6-Degrees-of-Freedom (DoF) rather than on the 2D image plane~\cite{Kehl2017, Rad2017, Tekin2018, Xiang2018, Manhardt_2018_ECCV, Li2018, Xu2018, Manhardt2019}.
This is motivated by a strong interest in achieving robust and accurate monocular 6D pose estimation for applications in the field of robotic grasping, scene understanding and augmented/mixed reality, where the use of a 3D sensor is not feasible~\cite{Pinto2016,Kokkinos2016,conf/cvpr/YaoFU12,8100178}.

Nevertheless, 6D pose estimation from RGB is a challenging problem due to the intrinsic ambiguity caused by visual appearance of objects under different viewpoints and occlusion. Indeed, most common objects exhibit shape ambiguities and repetitive patterns that cause their appearance to be very similar under different viewpoints, thus rendering pose estimation a problem with multiple correct solutions. Furthermore, also occlusion (from the same object or from others) can cause pose ambiguity. 

\begin{figure}[t!]
    \centering
    \includegraphics[width=0.975\linewidth]{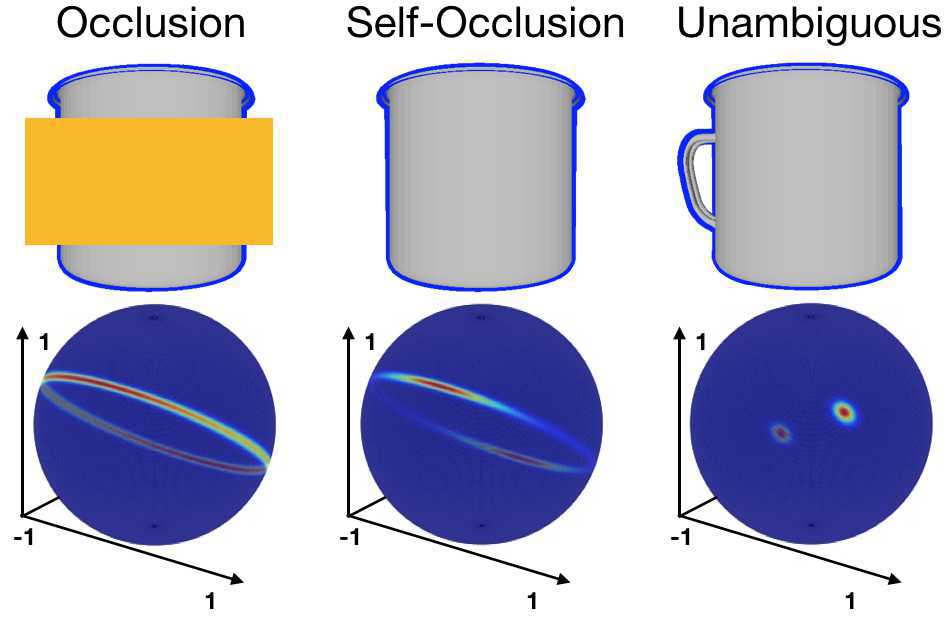}
    \caption{\textbf{Pose ambiguities}. External or self-occlusion can cause the 6DoF pose of an object to become ambiguous. Our method is able to detect and predict these ambiguities automatically without additional supervision. The antipodally symmetric Bingham distributions show that the model has understood the full range of valid poses.}
    \label{fig:ambig_multi_view}
\end{figure}

For example, as illustrated in Figure \ref{fig:ambig_multi_view}, the cup is identical from every viewpoint in which the handle is not visible. Thus, from a single image, it is impossible to univocally estimate the current object pose. Moreover, object symmetry can also induce visual ambiguities leading to multiple poses with the same visual appearance. However, most datasets do not reflect this ambiguity, as the ground truth pose annotations are mostly uniquely defined at each frame. This is problematic for a proper optimization of the rotation, since a visually correct pose still results in a high loss. Thus, many recent 3D detectors avoid regressing the rotation directly and, instead, explicitly model the solution space in an unambiguous fashion \cite{Rad2017,Kehl2017}.

Essentially, in ~\cite{Kehl2017}, the authors train their convolutional neural network (CNN) by mapping all possible pose solutions for a certain viewpoint onto an unambiguous arc on the view sphere. Rad \etal~\cite{Rad2017} employ a separate CNN solely trained to classify the symmetry in order to resolve these ambiguities. However, this simplification exhibits several downsides, such as the explicit inclusion of information about certain symmetries in each trained object. Moreover, this is not always easy to model, as \eg in the case of partial view ambiguity. Further, all these approaches rely on prior knowledge and annotation of the object symmetries and aim to solve the ambiguity by providing a single outcome in terms of estimated pose and object. Added to this, these methods are also unable to deal with ambiguities generated by other common factors such as occlusion.

On the contrary,  Sundermeyer \etal~\cite{Sundermeyer_2018_ECCV} and Corona \etal~\cite{DBLP:conf/iros/CoronaKF18} recently proposed novel methods to conduct pose estimation in an ambiguity-free manner. In the core, both learn a feature embedding solely based on visual appearance. Nonetheless, although \cite{Sundermeyer_2018_ECCV} is able to deal with ambiguities implicitly, it does not model their detection and description explicitly. In contrast, \cite{DBLP:conf/iros/CoronaKF18} also learns to classify the order of rotational symmetry, in particular the number of equivalent views around an axis of rotation. However, they require explicit hand-annotated labels and, in addition, cannot deal with ambiguities aside from these symmetry classes such as (self-) occlusion.

\begin{figure}[t!]
    \centering
    \includegraphics[width=\linewidth]{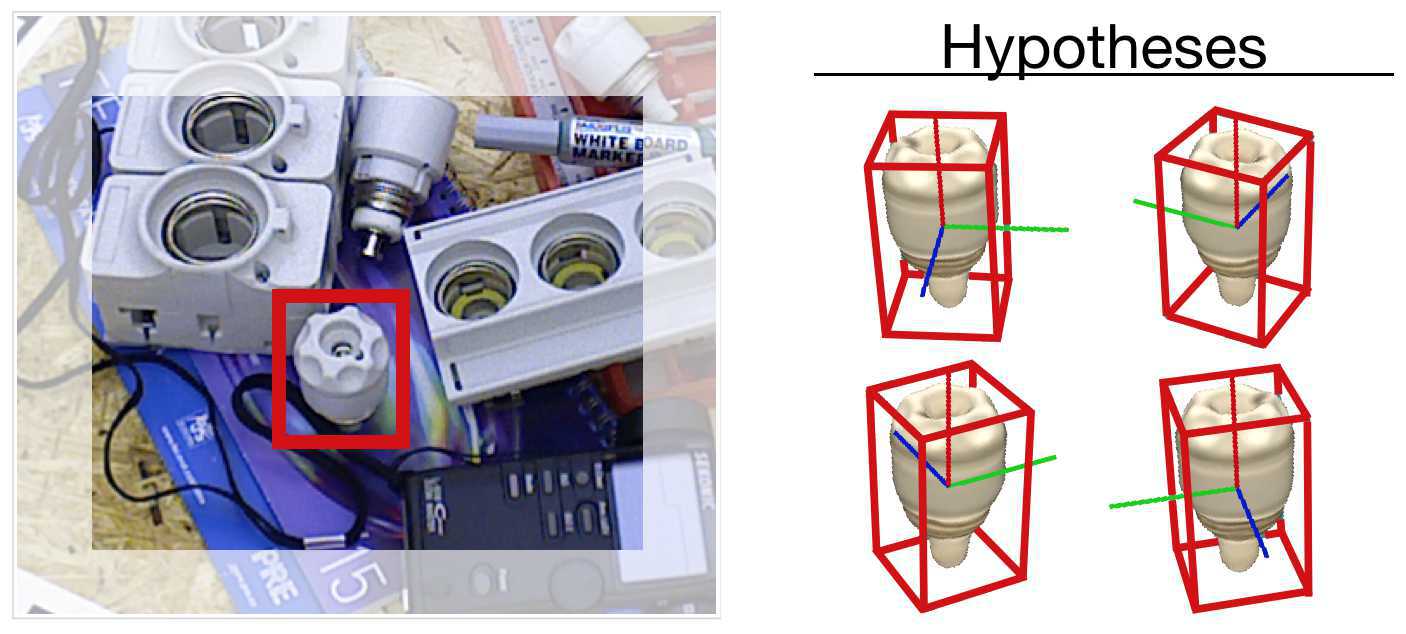}
    \caption{\textbf{Overview.} We predict $M$ hypotheses for the pose to approximate the distribution in the solution space. Each hypothesis is visually identical from the current viewpoint.}
    \label{fig:m_hypos}
\end{figure}

In this paper we propose to model the ambiguity of the object detection and pose estimation tasks directly by allowing our learned model to predict multiple solutions, or \emph{hypotheses}, for a given object's visual appearance (Fig~\ref{fig:m_hypos}). 
Inspired by Rupprecht \etal~\cite{rupprecht2017learning} we propose a novel architecture and loss function for monocular 6D pose estimation by means of multiple predictions.
Essentially, each predicted hypothesis itself corresponds to a 3D translation and rotation. When the visual appearance is ambiguous, the model predicts a point estimate of the distribution in 3D pose space. Conversely, when the object's appearance is unique, the hypotheses will collapse into the same solution. Importantly, our model is capable of learning the distribution of these 6D hypotheses from one single ground truth pose per sample, without further supervision.

Besides providing more insight and a better explanation for the task at hand, the additional knowledge gained from rotation distributions can be exploited to improve the accuracy of the pose estimates. In essence, analyzing the distribution of the hypotheses enables us to classify if the current perceived viewpoint is ambiguous and to compute the axis of ambiguity for that specific object and viewpoint. Subsequently, when ambiguity is detected, we can employ mean shift~\cite{Comaniciu02meanshift} clustering over the hypotheses in quaternion space to find the main modes for the current pose. A robust averaging in 3D rotation space for each mode then yields a highly accurate pose estimate. When the view is ambiguity-free, we can improve our pose estimates by robustly averaging over all 6D hypotheses, and by taking advantage of the predicted pose distribution as a confidence measure.

Our contributions are threefold:
\begin{itemize}
    \item We propose a novel method for 6DoF pose estimation, which can deal with the inherent ambiguities in pose by means of multiple hypotheses. 
    \item Explicit detection of rotational ambiguities and characterization of the uncertainty in the problem without further annotation or supervision. 
    \item A mechanism to measure the reliability and to increase the robustness of the unambiguous 6D pose prediction.
\end{itemize}
\section{Related Work}

We first review recent work in object detection and pose estimation from 2D and 3D data. Afterwards, we discuss common grounds and main differences with approaches aimed at symmetry detection for 3D shapes.

\paragraph{Object Detection and Pose Estimation.}

Almost all current research focus on deep learning-based methods. 
 
\cite{Wohlhart2015,Kehl2016a,DBLP:conf/iros/CoronaKF18} employ CNNs to learn an embedding space for the pose and class from RGB-D data, which can subsequently be utilized for retrieval. 
Notably, the majority of most recent deep learning based methods focus on RGB as input~\cite{Kehl2017,Rad2017,Do2018,Tekin2018,Xiang2018,Sundermeyer_2018_ECCV}. Since utilizing pre-trained networks often accelerates convergence and leads to better local minima, these methods are usually grounded on state-of-the-art backbones for 2D object detection, such as Inception~\cite{43022} or ResNet~\cite{He2016}. In particular, Kehl \etal~\cite{Kehl2017} employ SSD~\cite{Liu2016} with an InceptionV4~\cite{Szegedy2016} backbone and extend it to also classify viewpoint and in-plane rotation. Similarly, Sundermeyer \etal~\cite{Sundermeyer_2018_ECCV} also use SSD for localization, but employ an augmented auto-encoder for the unambiguous retrieval of the associated 6D pose. Rad \etal~\cite{Rad2017} utilize VGG~\cite{DBLP:journals/corr/SimonyanZ14a} and augment it to provide the 2D projections of the 3D bounding box corners. A similar approach is chosen by~\cite{Tekin2018}, based on YOLO~\cite{Redmon2016}. Afterwards, both apply P$n$P to fit the associated 3D bounding box into the regressed 2D projections, in order to estimate the 3D pose of the detection. In~\cite{Xiang2018}, Xiang \etal compute a shared feature embedding for subsequent object instance segmentation paired with pose estimation.

Finally, Do \etal~\cite{Do2018} extend Mask-RCNN \cite{He2017} with a third branch, which provides the 3D rotation and the distance to the camera for each prediction. 

\paragraph{Object Symmetry Detection}
Oftentimes, object pose ambiguity arises from symmetric shapes. We review relevant methods that extract symmetry from 3D models to outline commonalities and differences with our approach.

To our knowledge, \cite{DBLP:conf/iros/CoronaKF18} is the only method which estimates both: the 6D pose, and the symmetry of the perceived object. In particular, the network is trained to also predict the rotational order (\ie the number of identical views), posing it as a classification task.

Generally, most methods for symmetry detection are found in the shape analysis community. Among the different kinds of symmetries, axial symmetries are of particular interest, and multiple approaches have been proposed. Most methods rely on feature matching or spectral analysis: \cite{Eaton2009} treat the problem as a correspondence matching task between a series of keypoints on an object, determining the reflection symmetry hyperplane as an optimization problem. Elawady \etal~\cite{Elawady2017} rely on edge features extracted using a Log-Gabor filter in different scales and orientations coupled with a voting procedure on the computed histogram of local texture and color information. In addition, \cite{Cicconet2017} and \cite{Ovsjanikov2008} are also grounded on wavelet-based approaches. Recently, neural network approaches have also been proposed. Ke \etal~\cite{Ke2017} adapt an edge-detection architecture with multiple residual units and successfully apply it to symmetry detection using real-world images.

Notably, all these approaches aim at detecting symmetries of 3D shapes alone, while our focus is to model the ambiguity arising from objects under specific viewpoints with the goal of improving and explaining pose estimation. 

\section{Methodology}
In this section we describe our method for handling symmetries and other ambiguities for object detection and pose estimation in detail. We will first define what we understand as an ambiguity. 

\subsection{Ambiguity in Object Detection and Pose Estimation}

We describe the rigid body transformations $SE\left(3\right)$ via the semi-direct product of $SO\left(3\right)$ and $\mathbb{R}^3$.
While for the latter, we use Euclidean 3-vectors, the algebra $\mathbb{H}_1$ of unit quaternions is used to model the spatial rotations in $SO\left(3\right)$.
A quaternion is given by
\begin{equation}\label{eq:quaternion}
\textbf{q}
		= q_1 \textbf{1} + q_2 \textbf{i} + q_3 \textbf{j} + q_4 \textbf{k}
    = \left(q_1, q_2, q_3, q_4\right),
\end{equation}
with $\left(q_1, q_2, q_3, q_4\right) \in \mathbb{R}^4$ and $\textbf{i}^2 = \textbf{j}^2 = \textbf{k}^2 = \textbf{i}\textbf{j}\textbf{k} = - \textbf{1}$.
We regress quaternions above the $q_1 = 0$ hyperplane and, thus, omit the southern hemisphere, such that any possible 3D rotation can be expressed by only one single quaternion.

Under ambiguities, a direct naive regression of the rotation as a quaternion will lead to poor results, as the network will learn to predict a rotation that is closest to all results in the symmetry group. This prediction can be seen as the (conditional) mean rotation. More formally, in a typical supervised setting we associate images $I_i$ with poses $p_i$ in a dataset $(I_i, p_i)$ where $i\in\{1,\ldots,N\}$. To describe symmetries, we define for a given image $I_i$, the set $\mathcal{S}(I_i)$ of poses $p$ that all have an identical image 

\begin{equation} \label{eq:symmetryset}
\mathcal{S}(I_i) = \left\{p_j | I_j = I_i \right\}.
\end{equation}

Note that in the case of non-discrete symmetries the set $\mathcal{S}$ will contain infinitely many poses, which in turn transforms the sums of $S$ in the following to integrals. For the sake of a simpler notation and a finite training set in practice, we chose to continue with a notion of a finite $|S|$.
The naive model $f(I, \theta)$, that directly regresses a pose $p'$ from $I$, optimizes a loss $\mathcal{L}(p, p')$ by minimizing

\begin{equation} \label{eq:naive_loss}
\theta^* = \argmin_\theta \sum_{i=1}^N \mathcal{L}(f_\theta(I_i), p_i)
\end{equation}

\noindent over the training set. However, due to symmetry, the mapping from $I$ to $p$ is not well defined and cannot be modeled as a function. By minimizing Equation \ref{eq:naive_loss}, $f$ is learned to predict a pose $\tilde{p}$ approximating all possible poses for this image equally well.

\begin{equation}
f(I_i, \theta^*) = \tilde{p} =\min_p \sum_{j=1}^{| \mathcal{S}(I_i) |} \mathcal{L}(p, p_j)
\end{equation}

This is an unfavorable result since $\tilde{p}$ is chosen to minimize the sum of all losses towards the different symmetries. In the following section, we will describe how we model these ambiguities inside our method.

\subsubsection{Multiple Pose Hypotheses}

The key idea behind the proposed method is to model the ambiguity by allowing multiple pose predictions from the network. In order to predict $M$ pose hypotheses from $f$, we extend the notation to $f_\theta(I) = (f^{(1)}_\theta(I), \ldots, f^{(M)}_\theta(I))$ where $f$ now returns $M$ pose hypotheses for each image $I$.

For training, the idea is not to punish all hypotheses given the current pose annotation, since they might be correct under ambiguities. Thus, we use a loss that optimizes only one of the $M$ hypotheses for each annotation. The most intuitive choice is to pick the closest one. We adapt the meta loss $\mathcal{M}$ from \cite{rupprecht2017learning} that operates on $f$,

\begin{equation} \label{eq:meta_loss}
\theta^* = \argmin_\theta \sum_{i=1}^N \mathcal{M}(f_\theta(I_i), p_i),
\end{equation}

\noindent while we use the original pose loss $\mathcal{L}$ for each $f^{(j)}$

\begin{equation} \label{eq:mhp_loss}
\hat{\mathcal{M}}(f_\theta(I), p) = \min_{j=1,\ldots,M} \mathcal{L}(f_\theta^{(j)}(I), p).
\end{equation}

However, the hard selection of the minimum in equation \ref{eq:mhp_loss} does not work in practice as some of the hypothesis functions $f_\theta^{(j)}(I)$ might never be updated if they are initialized far from the target values. We relax $\hat{\mathcal{M}}$ to $\mathcal{M}$ by adding the average error for all hypotheses with an epsilon weight:

\begin{equation}\label{eq:m_loss}
  \begin{aligned}
        \mathcal{M}(f_\theta(I), p) = &
        \left(1-\epsilon\frac{M}{M-1}\right) \hat{\mathcal{M}}(f_\theta(I), p) \hspace{1mm} +\\    
        & \frac{\epsilon}{M-1}\sum_{j=1}^{M} \mathcal{L}(f_\theta^{(j)}(I), p).
    \end{aligned}
\end{equation}

\noindent The normalization constants before the two components are designed to give a weight of $(1-\epsilon)$ to $\hat{\mathcal{M}}$ and $\epsilon$ to the gradient distributed over all other hypotheses. When $\epsilon \rightarrow 0$, $\mathcal{M} \rightarrow \hat{\mathcal{M}}$. This is necessary since the average in the second term already contains the minimum from the first one.

\subsection{Architecture}
We employ SSD-300~\cite{Liu2016} with an extended InceptionV4~\cite{Szegedy2016} backbone and adjust it to also provide the 6D pose along with each detection. In particular, we append two more 'Reduction-B' blocks to the backbone. Essentially, we branch off after each dimensionality reduction block and place in total $6.099$ anchor boxes to cover objects at different scales. Moreover, to include the unambiguous regression of the 6D pose, we modify the prediction kernel such that it provides $C + M \cdot P$ outputs for each anchor box. Thereby, $C$ denotes the number of classes, $M$ denotes the number of hypotheses, and $P$ denotes the number of parameters to describe the 6D pose. In our case, for each of the $M$ predicted hypotheses, we regress $P=5$ values to characterize the 6D pose, composed of an explicitly normalized 4D quaternion for the 3D rotation and the object's distance towards the camera. We can estimate the remaining two degrees-of-freedom by back-projecting the center of the 2D bounding box using the inferred depth.

Additionally, in line with \cite{Liu2015, Kehl2017} we conduct hard negative mining to deal with foreground-background imbalances. Thus, given a set of positive boxes \textit{Pos} and hard-mined negative boxes \textit{Neg} for a training image, we minimize the following energy function:

\begin{equation}\label{eq:loss}
\begin{split}
        \mathcal{L}(\Pos, \Neg) := \sum_{b\in \Neg}\mathcal{L}_{class} + \\
        \sum_{b\in \Pos} (\mathcal{L}_{class} + \alpha \mathcal{L}_{fit} + \beta \mathcal{M}(f_\theta(I), p)).
\end{split}
\end{equation}

\noindent For the class and the refinement of the anchor boxes, we employ the cross-entropy loss $\mathcal{L}_{class}$ and the smooth L1-norm $\mathcal{L}_{fit}$, respectively.
In order to compare the similarity of two quaternions, we compute the angle between the estimated rotation and the ground truth rotation according to

\begin{equation}
   \mathcal{L}_{rotation}(q, q') = \arccos \left( 2 \langle q, q'\rangle^2 - 1 \right).
\end{equation}

\noindent Additionally, we employ the smooth L1-norm as loss for the depth component $\mathcal{L}_{depth}$.   

Altogether, we define the final loss for each hypothesis $j$ and input image $I$ as follows
\begin{equation}\label{eq:loss_final}
   \mathcal{L}(f_\theta^{(j)}(I)) = \mathcal{L}_{rotation}(q^{(j)},q') + \lambda\mathcal{L}_{depth}(d^{(j)},d').
\end{equation}
\subsection{Processing Multiple Hypotheses}
\label{post_processing}

During inference we further analyze the predicted multiple hypotheses in order to determine whether the pose of the object is ambiguous. Notice that prior to this, we first map all hypotheses to reside on the upper hemisphere. If we detect an ambiguity, we additionally exploit the multiple hypotheses to estimate the view-dependent axes of ambiguity.

\paragraph{Detection of Visual Ambiguities in Scenes.}

We analyze the distribution of predicted hypotheses in quaternion space to determine whether the pose exhibits an ambiguity. To this end, Principal Component Analysis (PCA) is performed on the quaternion hypotheses $\textbf{q}_i$. The singular value decomposition of the data matrix indicates the ambiguity: if the dominant singular values $\sigma_{1/2} \gg 0$ ($\sigma_i > \sigma_{i+1} \ \forall i$), an ambiguity in the pose prediction is likely, while small singular values imply a collapse to a single unambiguous solution.

We determine the existence of ambiguity by thresholding the value of $\sigma_2$. Empirically, we find the criteria $\sigma_2 > 0.8$ to offer good estimations for ambiguity. 
It is noteworthy that we can learn to detect ambiguities without further supervision, directly from standard datasets. 

\begin{figure}[t!]
	\centering
	\hspace{-4mm}
	\includegraphics[width=0.9\linewidth]{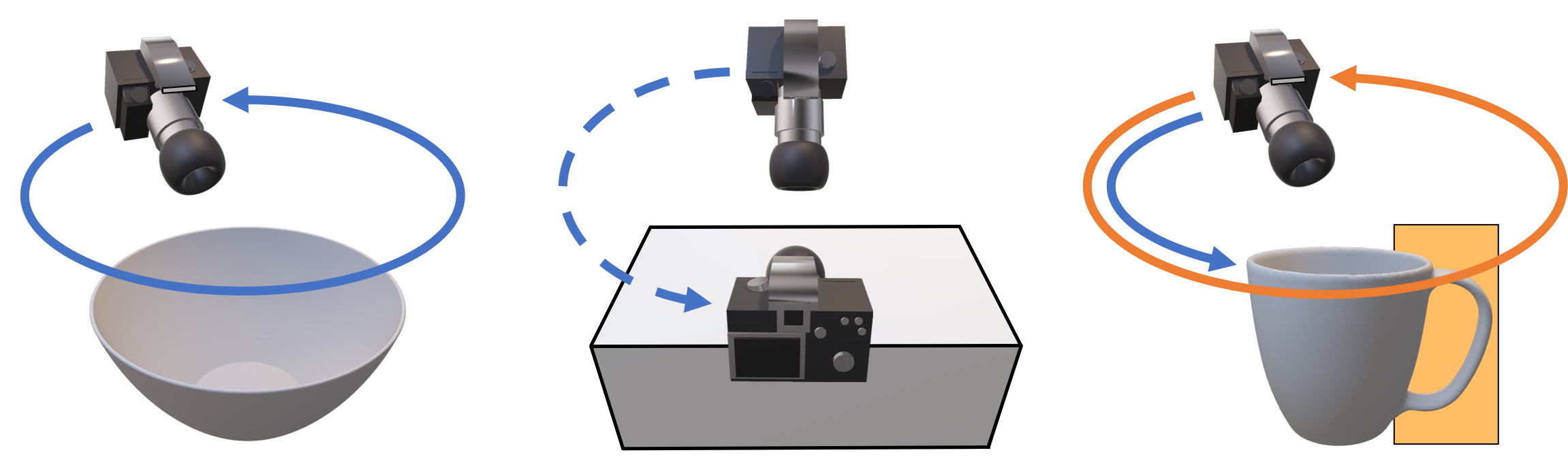}
	\caption{\textbf{Examples of pose ambiguity.} Left: Rotational ambiguity. Mid: Two different possible poses for each side. Right: Ambiguity around an arc through (self-) occlusion.}
	\label{fig:symmetry_examples}
\end{figure}

\paragraph{Estimation of the Axis of Ambiguity.}
As mentioned, very prominent representatives for visual ambiguities are symmetries in the objects of interest, as illustrated in Fig.~\ref{fig:symmetry_examples} (left) and (mid).
Nevertheless, for other objects such as cups, also (self-) occlusion can induce ambiguities in appearance (right).

To calculate a viewpoint dependant ambiguity axis, we take a closer look at the following scenario.
A rotation $\textbf{q}_i = \left(q_{i1}, q_{i2}, q_{i3}, q_{i4}\right)$ rotates the camera $c_0$ to $c_i$ around the rotation axis
\begin{equation}
a_i = \left( q_{i2}, q_{i3}, q_{i4} \right) /  \sqrt{q_{i2}^2 + q_{i3}^2 + q_{i4}^2}.
\end{equation}
All these rotation axes lie in the same plane which is perpendicular to the ambiguity axis $s \perp a_i \ \forall i$.
Thus, if we stack the rotation axes
$A = \left( a_1^T, a_2^T, \cdots, a_n^T \right)$,
we can formulate the overdetermined linear equation system $A^T s = 0$. The ambiguity axis can be found as the solution to the optimization problem
\begin{equation}
\min_{s \in \mathbb{R}^3} \left\lVert A^T s \right\rVert_p,
\end{equation}
which we solve for $p=2$ using SVD. 

\subsection{From Multiple Hypotheses to 6D Pose}\label{6dpose}\label{robust_pose}
After analyzing the distribution of the hypotheses, we can robustly compute the associated 6D pose for each case.

\paragraph{Unambiguous Object Pose.}
In case of an unambiguous object pose, we utilize the multiple hypotheses as an input for a geometric median (geodesic $L_1$-mean~\cite{hartley2013}) to improve robustness of the overall estimation
\begin{equation} \label{eq:geometricmedian}
\textbf{q}_{\text{gm}} = \argmin_{\textbf{q} \in \mathbb{H}_1} \sum_i \text{d}_{\text{geo}} \left( \textbf{q}_i, \textbf{q} \right).
\end{equation}

\noindent The iterative calculation follows the Weiszfeld algorithm~\cite{weiszfeld1937,hartley2011l1} in the tangent spaces to the quaternion hypersphere~\cite{busam2017}.
From a statistical perspective, our rotation measures are treated as inputs for an $L_1$-estimator to robustly detect the geometric median where $\text{d}_{\text{geo}}$ gives the geodesic distance on the quaternion hypersphere. Note that Gramkow~\cite{gramkow2001averaging} showed that locally, using the Euclidean distance in the ambient, quaternion space well approximates the Riemannian one. In addition, we compute the median depth of all hypotheses. Afterwards, we utilize the center of the 2D detection and backproject it into 3D to obtain the translation and therewith the full 6D pose of the detection.

\paragraph{Ambiguous Object Pose.}\label{clustering}
As the number of possible 3D rotations is finite yet unknown, we employ mean shift~\cite{Comaniciu02meanshift} to cluster the hypotheses in quaternion space. 
Specifically, we use the the angular distance of the quaternion vectors to measure similarity and the Weiszfeld algorithm to merge clusters inside mean shift. 
This yields either one cluster (if the poses are connected) or multiple (if they are unconnected) as illustrated in Fig.~\ref{fig:symmetry_examples}. 
For each cluster we compute a median rotation and the median depth to retrieve the associated 3D translation. Note that we only consider the depths of the hypotheses, which contributed to the corresponding cluster. We apply simple contour checks \cite{Kehl2017} to find the best fitting cluster from which we extract the final 6D pose.

\paragraph{Synthetic Data.}
As noted in \cite{Hinterstoisser2017}, domain adaptation between synthetically generated data samples and real-world images trivializes the collection of training data. We render CAD models in random poses and add a series of augmentations, such as illumination changes, shadows and blur, as well as background images taken from the MS COCO~\cite{Lin2014a}.

\begin{figure*}[t!]
	\centering
	\includegraphics[width=.965\linewidth]{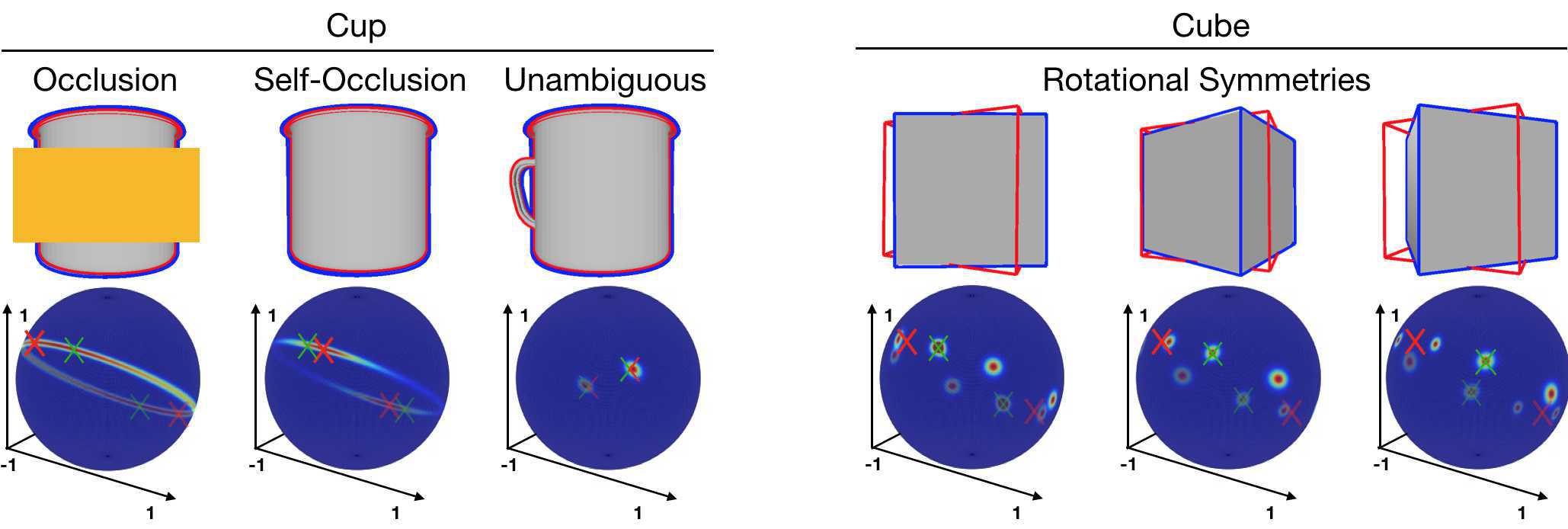}
	\caption{\textbf{Synthetic toy dataset.} Top: Contours of the rendered poses for the naive $\shp$ (M=1) model in red and our $\mhp$ (M=30) model in blue. Bottom: Bingham distributions for each pose cluster, together with the ground truth quaternion in green and the $\shp$ predicted quaternion in red. Our model is not only correct in both cases but can also predict the full range of valid poses. $\shp$ fails on the cube example.}
	\label{fig:toy_pose_bingham}
\end{figure*}

\section{Evaluation}
\label{sec_experiments}

In this section, we first introduce our experimental setup. Following that, we clearly demonstrate the benefits of our method compared to typical pose estimation systems on a toy dataset. Next, we show robustness in determining whether a view exhibits an ambiguity. Fourth, we report our 6D pose estimation accuracy for the unambiguous and the ambiguous case on common benchmark datasets. Finally, we demonstrate how we can model reliability in pose estimation by analyzing the variance across hypotheses. 

\subsection{Experimental Setup}

\paragraph{Evaluation metrics.}

In order to properly assess the 6D pose performance, we distinguish between potentially ambiguous and non-ambiguous objects.
When dealing with non-ambiguous objects, we report the absolute error for the 3D rotation in degrees and 3D translation in millimeters. We also show our accuracy using the Average Distance of Distinguishable Model Points (ADD) metric from \cite{Hinterstoisser2012}, which measures if the average deviation of the transformed model points is less than $10\%$ of the object's diameter. 

For `ambiguous' objects we rely on the Average Distance of Indistinguishable Model Points (ADI) metric, which extends ADD for ambiguity, measuring error as the average distance to the \emph{closest} model point \cite{Hodan2016, Hinterstoisser2011}.

We also show our results for the Visual Surface Similarity (VSS) metric. As \cite{Kehl2017}, we define VSS similar to the Visual Surface Discrepancy (VSD) \cite{Hodan2016}, however, set $\tau=\infty$. Hence, we measure the pixel-wise overlap of the rendered ground truth pose and the rendered prediction, which is not subject to ambiguities.

\paragraph{Bingham Distributions.}

In order to visually analyze the multi-hypotheses output of our network, we inspect the underlying rotation distributions. A \textit{Bingham distribution}~\cite{bingham1974antipodally} (BD) is a special equivalent to a Gaussian distribution on a hypersphere. BDs represent a probability distribution on $S^d$ with antipodal symmetry well suited to study poses parametrized by quaternions, where $q$ and $-q \in \mathbb{H}_1$ represent the same element in $SO\left(3\right)$.
In line with previous works~\cite{kurz2013recursive, glover2014tracking, birdal2018bayesian}, we visualize an equatorial projection of the closest distribution to our pose output using BDs.

\subsection{Synthetic Ambiguity Evaluation}
\begin{table}[t!]
\centering
\scalebox{0.79}{
    \begin{tabular}{@{}c|c|c|c|c|c@{}}
        \multirow{ 2}{*}{Object} & \multirow{2}{*}{Ambiguity} &\multicolumn{2}{c}{$\shp$} & \multicolumn{2}{|c}{$\mhp$} \\
        &  & VSS [\%] & ADI [\%] & VSS [\%] & ADI [\%] \\
        \midrule
        Cup & (Self-) Occlusion & 97.0 & \textbf{100} & \textbf{98.1} & \textbf{100} \\
        Cube & Plane Symmetries & 87.4 & 15.6 & \textbf{98.6} & \textbf{100}
    \end{tabular}
    }
    \caption{\textbf{Synthetic results.} for the naive $\shp$ $(M=1)$ and our $\mhp$ ($M=30$) model on the synthetic toy dataset.}
    \label{tab:toy_sample}
\end{table}
We render a simple synthetic dataset of a rotating cup and cube. We compare the baseline with $M=1$ hypothesis and our method with $M=30$ hypotheses. The results are shown in Fig.~\ref{fig:toy_pose_bingham}, Tab.\ref{tab:toy_sample}, and the supplement. For the cup, both methods yield an ADI score of 100\%. The single hypothesis approach $\shp$ is indeed able to compute visually correct poses even though it cannot model the pose distribution along an arc. It has learned the conditional mean pose where the handle is exactly opposite of the camera. Nonetheless, this is only one of the infinitely many possible solutions. In contrast, our method is able to predict the whole distribution as seen in the Bingham plots. This is essential for tasks such as next-best-view prediction or robotic manipulation. When there is no ambiguity, both methods predict only the one correct pose.

For the cube object, $\shp$ fails (red outline) with an ADI of only 15.6\%. Here, the conditional mean is not inside the set of correct poses. Our method is again able to estimate the underlying distribution and can correctly estimate all four modes of correct poses. This yields a perfect ADI of 100\%. 

\begin{figure}[t!]
	\centering
	\includegraphics[width=\linewidth]{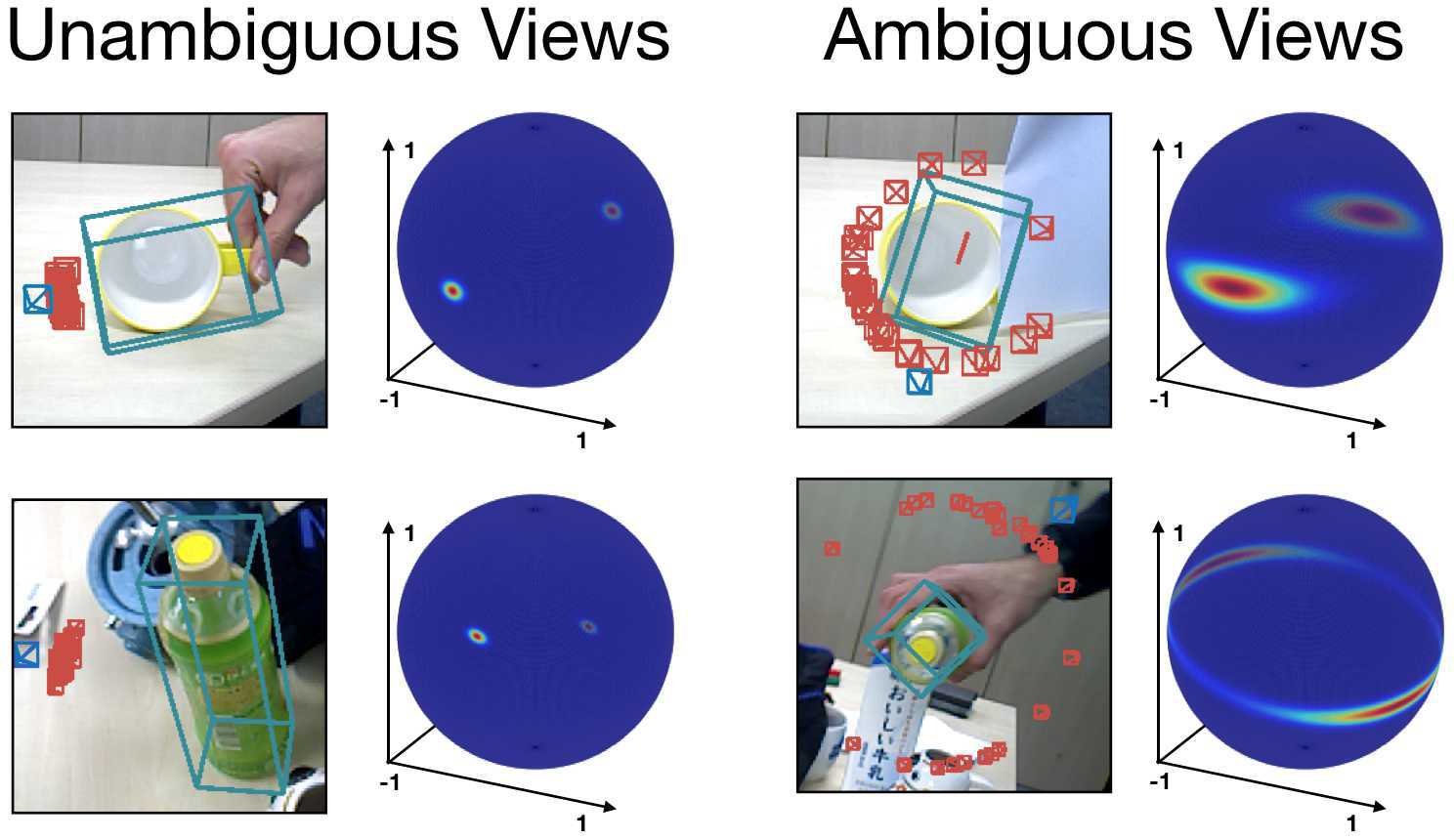}
	\caption{\textbf{Real data.} The red frustums visualize $(M=30)$ pose hypotheses. The blue frustum constitutes the median, which determines the predicted 3D bounding box. In the unambiguous case (left) the hypotheses agree. However, partial symmetries and occlusion lead to multiple possible outcomes on the right, which meaningfully reflect to the Bingham distribution of hypotheses.}
	\label{fig:qual_pose}
\end{figure}

When applying our method to real data (Fig.~\ref{fig:qual_pose}), we achieve similar results. If there is a unique solution, the method is able to robustly estimate the correct pose. For ambiguous views, we retrieve the governing distribution as depicted by the viewpoint frustums and spherical plots.

\subsection{Real World Datasets}
To conduct evaluations on real data, we build two datasets addressing both \textit{unambiguous} and \textit{ambiguous} cases.
In particular, for the former, we use the popular `LineMOD' \cite{Hinterstoisser2012} and `LineMOD Occlusion' dataset~\cite{Krull2015}. The  authors of \cite{Krull2015} selected one sequence from the original `LineMOD' dataset and  labeled eight additional objects. Nevertheless, we moved the `glue' and `eggbox' object to the \textit{ambiguous dataset}, since both exhibit several views (mostly from the top), which are not unique. Additionally, following \cite{Kehl2017, Rad2017} we removed the `cup' and `bowl' objects, because no watertight CAD models are provided for them. We also discard the `lamp' since the CAD model does not possess correct normal vectors for proper rendering. To the latter, the \textit{ambiguous} dataset, besides the `glue' and `bowl' objects, we added several models from T-LESS~\cite{Hodan2017} to cover different types of ambiguities. In essence, T-LESS mostly consists of symmetric and textureless industrial objects. For our experiments we choose a subset that covers both cases: complete rotational symmetry along an axis (object 4) and objects with more than one rotational symmetry (object 5, 9, 10).

\subsection{Ambiguity Detection Analysis}
\begin{figure}[t]
    \centering
    \includegraphics[width=0.4875\linewidth]{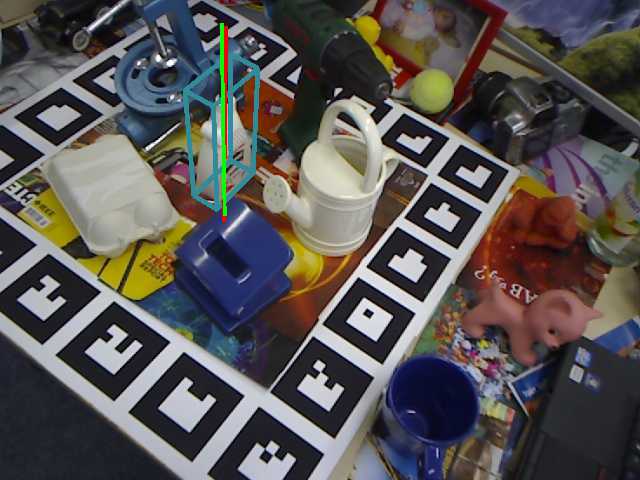}
    \includegraphics[width=0.4875\linewidth]{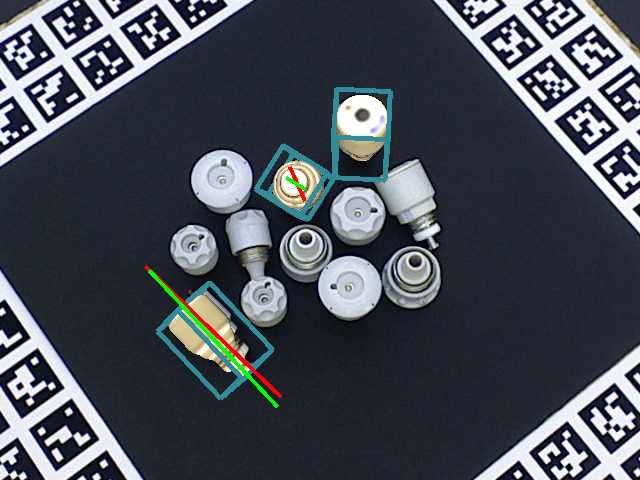}
    \caption{\textbf{Ambiguity detection.} Symmetry axis (green line) estimation\label{fig:amg_detection}. Notice that one screw was classified to be unambiguous (\ie no axis), because the ambiguity could be resolved through the texture.}
\end{figure}
To evaluate the ability of our model to learn pose distributions, we manually labeled for each validation image of the \textit{ambiguous} dataset, whether the current object view exhibits ambiguity based on the visible object texture and shape.
This ground truth is used to quantitatively assess our capability of detecting pose ambiguity. Additionally, we compute the ground truth symmetry axis for each object. It is important to note that we do not conduct object symmetry detection, instead, we describe the perceived pose ambiguity in terms of a symmetry axis. These annotations are only used for evaluation and not during training.

For each detected ambiguity, we compute the average discrepancy of the computed symmetry axis from the ground truth annotation. For the ambiguity-free case, we achieve to report an accuracy of more than 99\%, while for the ambiguous case we can also state a high accuracy of 82\% correctly classified views. Furthermore, the mean axis only deviates by 24\degree, which shows that our formulation is able to precisely explain the perceived ambiguity.

In Fig.~\ref{fig:amg_detection}, we respectively show one sample of estimated ambiguity axis from `LineMOD' and `T-LESS'. For each detection, we draw the estimated axis in red, while the green line denotes the hand-annotated groundtruth axis.

\subsection{Comparison to State-of-the-Art}
\paragraph{Unambiguous Pose Estimation.}
In Tab~\ref{table:syn_linemod_pose} and Tab~\ref{table:real_linemod_pose}, we report our results for the \textit{unambiguous} subset for training with synthetic data and with the train data split from \cite{Brachmann2016}. Since the number of predicted hypotheses $\mathbf{M}$ is a hyperparameter, we will show an ablation in the supplement and only report our best results with $\mathbf{M=5}$ here.

\begin{table}[t!]
	\begin{center}
	\scalebox{0.79}{
		\begin{tabular}{@{}c||c|c|c|c||c@{}}
			& Rot. [\degree] & Trans. [mm] & VSS [\%]& ADD [\%] & F1 \\ \midrule
			SSD-6D~\cite{Kehl2017} & 28.0 & 72.4 & 67.4 & 9.4 & 88.8 \\
			\cite{Sundermeyer_2018_ECCV} & -- & -- & -- & 22.1 & -- \\
			\midrule
			$\shp \mathbf{(M = 1)}$ & 17.9 & 45.6 & 76.8 & 31.2 & 91.6 \\
			$\mhp \mathbf{(M = 5)}$ & \textbf{17.4} & \textbf{39.5} & \textbf{78.2} & \textbf{35.3} & \textbf{93.4} 
		\end{tabular}
      }
	\end{center}
	\caption{\textbf{Pose errors of unambiguous objects with synthetic training data}.
	Comparison with \cite{Sundermeyer_2018_ECCV}, \cite{Kehl2017}. Results of \cite{Kehl2017} from their released models and code.}
	\label{table:syn_linemod_pose}
\end{table}

For the case of synthetic training only, even for the single hypothesis case, our approach outperforms SSD-6D by more than 35\% of relative error while also being more robust in terms of 2D detection. Comparing with Sundermeyer \etal\cite{Sundermeyer_2018_ECCV} we can report a relative improvement of approximately 50\% referring to ADD. In addition, our averaging over all hypotheses leads to more robustness towards outliers and, thus, another improvement of all metrics.

\begin{table}[t!]
	\begin{center}
	\vspace{1mm}
	\scalebox{0.79}{
	\begin{tabular}{@{}c||c|c|c|c|c|c||c@{}}
		 & ape & can & cat & dril & duck & holep & mean\\ \midrule
		 Tekin~\cite{Tekin2018} & 2.5 & 17.5 & 0.7 & 7.7 & 1.1 & 5.5 & 5.8 \\
		 $\mhp \mathbf{(M = 5)}$  & \textbf{5.9} & \textbf{22.4}& \textbf{4.2} & \textbf{32.0} & \textbf{12.2} & \textbf{17.0} & \textbf{15.6}
    \vspace{2mm}
	\end{tabular}
	}
	\scalebox{0.79}{
	\begin{tabular}{@{}c||c|c|c}
    	& BB-8 \cite{Rad2017} & Tekin \cite{Tekin2018} & $\mhp \mathbf{(M = 5)}$  \\
    	\midrule
    	ADD [\%] & 45.9 & \textbf{47.9} & 44.4   
    	\end{tabular}
	}
		\caption{\textbf{Pose errors of unambiguous objects with real training data split from \cite{Brachmann2016}}. Top: Comparison with \cite{Tekin2018} on LineMOD Occlusion. Bottom: Comparison with \cite{Rad2017} and \cite{Tekin2018} on LineMOD. Results of \cite{Tekin2018} from their released models and code.}
		\label{table:real_linemod_pose}
	\end{center}
\end{table}
When also employing real data, we can improve our results by approximately 9\% to 44.4\% and are on par with the state-of-the-art methods from \cite{Rad2017} and \cite{Tekin2018}, even though we employ no crop and paste augmentations. Further, when using the more challenging `LineMOD Occlusion' dataset, we can exceed Tekin \etal \cite{Tekin2018} for all objects and overall almost triple their ADD score from 5.8\% to 15.6\%.

\paragraph{Ambiguous Pose Estimation.}
\begin{table}[t!]
    \begin{center}
     \scalebox{0.79}{
        \begin{tabular}{@{}c||c|c|c|c|c|c||c|c|c@{}}
             & \multicolumn{3}{c|}{VSS [\%]} & \multicolumn{3}{c||}{ADI [\%]} & \multicolumn{3}{c}{F1} \\
             & $\mhp$ & $\shp$ & \cite{Kehl2017} & $\mhp$ & $\shp$ & \cite{Kehl2017} & $\mhp$ & $\shp$ & \cite{Kehl2017} \\ \midrule
            eggbox & \textbf{83.1} & 78.5 & 76.3 & 55.7 & \textbf{56.0} & 26.3 & \textbf{98.0} & 83.0 & 93.7\\
            glue &  \textbf{74.6} & 74.0 & 65.1 & 54.6 & \textbf{58.7} & 17.6 & \textbf{90.1} & 74.0 & 76.8 \\ 
            \midrule
            \textbf{mean} & \textbf{78,9} & 76.3 & 70.7 & 55.2 &  \textbf{57.4} & 22.0 & \textbf{94.1} & 78.5 & 85.5
      \vspace{3mm}
        \end{tabular}
    }
     \scalebox{0.79}{
        \begin{tabular}{@{}c|c||c|c|c|c|c|c@{}}
            \multicolumn{2}{c||}{} & \multicolumn{3}{c|}{VSS [\%]} & \multicolumn{3}{c}{ADI [\%]} \\
            & Scene & $\mhp$ & $\shp$ & \cite{Sundermeyer_2018_ECCV} & $\mhp$ & $\shp$ & \cite{Sundermeyer_2018_ECCV}  \\ \midrule
            obj\_04 & 5, 9 & 70.8 & 68.6  & \textbf{78.5} & \textbf{19.7} & 14.1 & 15.2 \\
            obj\_05 & 2, 3, 4  & 87.6 & 82.8 & \textbf{88.8} & \textbf{78.0} & 48.3 & 76.3 \\
            obj\_09 & 5, 11  & 84.4 & 79.1 & \textbf{86.5} & 69.9 & 54.5 & \textbf{77.3} \\
            obj\_10 & 5, 11  & 82.0 & 78.5 & \textbf{82.3} &  \textbf{57.9} & 29.4 & 31.9 \\
            \midrule
            \multicolumn{2}{c||}{\textbf{mean}} & 81.2 & 77.3 & \textbf{84.0} & \textbf{56.4} & 36.6 & 50.6
        \end{tabular}
      }
    \caption{\textbf{Ambiguous dataset.} (top: `LineMOD') (bottom: T-LESS). We compare our multiple hypotheses $\mhp$ ($M = 30$) results and the same predictor trained to output a single hypothesis $\shp$ ($M = 1$) with \cite{Sundermeyer_2018_ECCV} and SSD-6D~\cite{Kehl2017}. 
   }
    \label{tab:linemod_ambig_pose}
    \label{tab:tless_individual_pose}
    \end{center}
\end{table}
Referring to Tab~\ref{tab:linemod_ambig_pose}, for the ambiguous `LineMOD' objects, we attain a VSS score of 79\% and an ADI score of 55\%, which is a relative improvement of approximately 13\% and 145\% compared to SSD-6D. In the 6D setting, the multiple hypothesis detector overall achieves similar performance as the single hypothesis predictor. However, for the 2D detection case, we are able to increase the accuracy from 79\% to 94\%. As constituted, only a few views are ambiguous for these objects. Investigating the results, we discovered that the single hypothesis predictor is not able to understand exactly these views and tends to simply discard them. In contrast, the multiple hypotheses predictor is indeed able to understand these views and yields reliable pose predictions.

For all ambiguous `T-LESS' objects (Tab~\ref{tab:tless_individual_pose}), our multiple hypotheses approach surpasses the single hypothesis estimator, which, when trained and evaluated under the same conditions, is not able to capture the ambiguities in pose. Thus, the single hypothesis predictor is not able to produce equally accurate results, being only capable of computing precise poses for unambiguous views.
Comparing with \cite{Sundermeyer_2018_ECCV}, we report similar performance in pose. Our ADI improves with $56.4\%$ compared to $50.6\%$ while VSS falls slightly behind by $2.8\%$. For fairness, we only compare the 6D pose accuracy for correctly detected objects (\ie $IoU\le0.5$) since \cite{Sundermeyer_2018_ECCV} trained their 2D detector for T-LESS on real data.

\subsection{Measuring Reliability}

\begin{figure}[t!]
    \scalebox{0.79}{
    	\begin{tabular}{@{}c|c|c|c|c|c@{}}
    		STD $\sigma$ & Rot. [\degree] & Trans. [mm] & VSS [\%] & ADD [\%] & Rejects [\%]\\ \toprule
    		$< 0.05$ & 11.8  & 39.4 & 80.0 & 37.7 & 32.6 \\
    		$< 0.075$ & 13.8 & 41.3 & 79.1 & 35.5 & 18.2 \\
    		$< 0.10$ & 15.5 & 43.0 & 78.3 & 34.3 & 10.5 \\
    		$< 0.15$ & 17.3 & 44.0 & 77.7 & 33.4 & 4.0 \\
    		$< \infty$ & 19.2 & 44.8 & 77.3 & 32.7 & 0.0 
            \vspace{2mm}
    	 \end{tabular}
      }
    \centering
    \includegraphics[width=0.4875\linewidth]{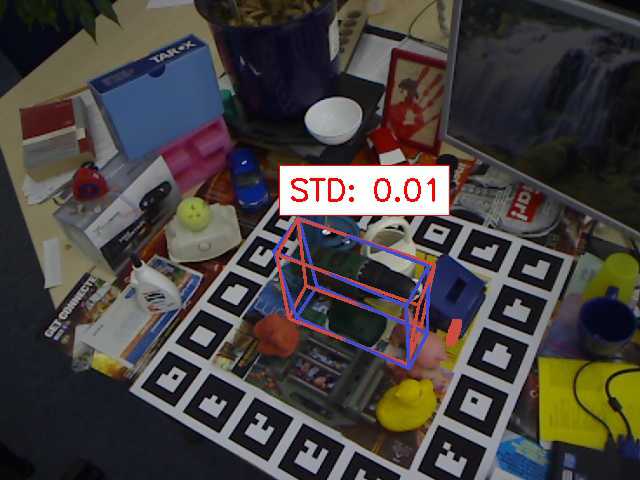}
    \includegraphics[width=0.4875\linewidth]{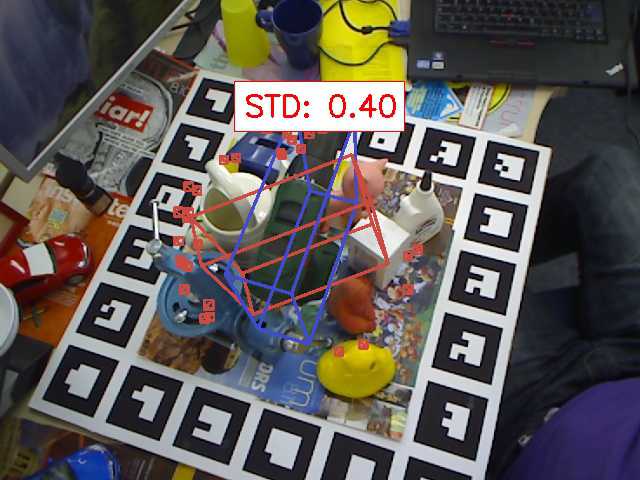}
      \caption{\textbf{Reliability.} Top: results for different bins for the standard deviation over all hypotheses for the poses. Bottom: pose with the lowest (left) and the highest (right) standard deviation in the hypotheses. GT pose in blue, predicted pose in red. The red frustums illustrate the hypotheses.
      \label{table:pose_confidence}}
\end{figure}

To the best of our knowledge, there is no prior work capable of 
modelling the confidence in the continuous pose estimate. Yet, this information can highly improve the overall robustness and accuracy. 
In our case, we can utilize the different hypotheses to first determine whether the current view is unambiguous and subsequently employ them as a confidence measurement in the unambiguous 6D pose. To quantify the effect of this, we report our test results on the unambiguous subset of `LineMOD' in Fig.~\ref{table:pose_confidence} (top), where we compute a confidence measure via the standard deviation with respect to the Karcher mean~\cite{karcher1977}. 
 
Naturally, a lower standard deviation means more accurate poses. By only allowing poses with $\sigma <0.1$, all metrics improve, while only losing about $10.5\%$ of all estimates. The rotational error decreases by approximately $20\%$ and the translation error drops from 44.8mm to 43.0mm. Accordingly, using an even lower threshold (\eg $\sigma < 0.05$) gives another significant improvement for pose (especially in rotation), however, at the cost of rejecting more estimates. 
The qualitative example image in Fig.~\ref{table:pose_confidence} also confirms these results. The pose with the lowest standard deviation for the `driller' is very accurate, and the one with the highest is rather imprecise. We experience the same behavior for all \textit{unambiguous} `LineMOD' objects.

\section{Conclusion}
 
We propose a new approach for pose estimation that implicitly models ambiguities without requiring any input pre-processing as well as the feasibility of domain adaptation between synthetic and real data. In addition, we can estimate the axis of rotational ambiguity and perform pose refinement based on clustering without knowing the number of clusters in advance. Our experiments show that our method is suitable for detecting both challenging objects with multiple rotational symmetries and datasets with little ambiguity.
Lastly, we argue that our method constitutes a metric of reliability for the 6D pose.

In conclusion, we believe that the new formulation of the pose detection problem from images as an ambiguous task paves the way towards interesting applications in the domain of robotic interactions and automation.

\paragraph{Acknowledgments} We would like to thank Toyota Motor Corporation for funding and supporting this work and NVIDIA for the donation of a GPU.

\newpage

\setcounter{section}{0}
\setcounter{figure}{0}
\setcounter{table}{0}
\onecolumn

\begin{center}
	\Large \textbf{Explaining the Ambiguity of Object Detection and 6D Pose from Visual Data}
\end{center}
\vspace{3.5mm}
\begin{center}
	\Large \textbf{Supplementary Material}
\end{center}
\vspace{10mm}

\begin{abstract}
    This document supplements our main paper entitled \textit{Explaining the Ambiguity of Object Detection and 6D Pose From Visual Data} by providing 1. details on the used datasets, 2. quantitative results for the quality of ambiguity prediction, 3. further quantitative and qualitative evaluations for the tasks of both unambiguous (single-hypothesis) pose estimation, and ambiguity characterization, 4. examples of confidence estimation, 5. Implementation details and pseudocode of our inference and finally 6. representative images and details of the synthetic training dataset we used to train our networks.
\end{abstract}

\begin{figure*}[!ht ]
    \centering
    	\subfloat[`ape']{\includegraphics[width=.06\textwidth]{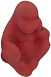}}
    	\hspace{1mm}
    	\subfloat[`bvise']{\includegraphics[width=.1\textwidth]{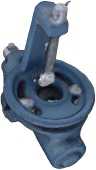}}
    	\hspace{1mm}
    	\subfloat[`cam']{\includegraphics[width=.1\textwidth]{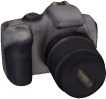}}
    	\hspace{1mm}
    	\subfloat[`can']{\includegraphics[width=.1\textwidth]{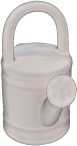}}
    	\hspace{1mm}
    	\subfloat[`driller']{\includegraphics[width=.1\textwidth]{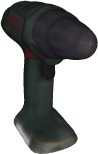}}
    	\hspace{1mm}
    	\subfloat[`duck']{\includegraphics[width=.075\textwidth]{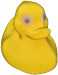}}
    	\hspace{1mm}
    	\subfloat[`holep']{\includegraphics[width=.1\textwidth]{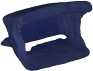}}
    	\hspace{1mm}
    	\subfloat[`iron']{\includegraphics[width=.1\textwidth]{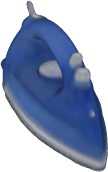}}
    	\hspace{1mm}
    	\subfloat['phone']{\includegraphics[width=.1\textwidth]{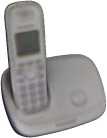}}
    	\\
        \setcounter{subfigure}{0}
        
    	\subfloat[`eggb']{\includegraphics[width=.1\textwidth]{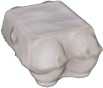}}
    	\hspace{10mm}
    	\subfloat[`glue']{\includegraphics[width=.07\textwidth]{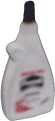}}
    	\hspace{10mm}
    	\subfloat[`obj\_04']{\hspace*{5pt} \includegraphics[width=.05\textwidth]{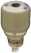}\hspace*{5pt}}
    	\hspace{10mm}
    	\subfloat[`obj\_05']{\includegraphics[width=.09\textwidth]{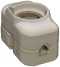}}
    	\hspace{10mm}
    	\subfloat[`obj\_09']{\includegraphics[width=.145\textwidth]{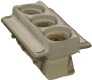}}
    	\hspace{10mm}
    	\subfloat[`obj\_10']{\includegraphics[width=.1\textwidth]{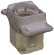}}
    	\caption{Top: 3D Models of the \textit{unambiguous} Dataset from `LineMOD'~\cite{Hinterstoisser2011}. Bottom: 3D Models of the \textit{ambiguous} Datasets from `LineMOD~\cite{Hinterstoisser2011} (first two) and T-Less~\cite{Hodan2017} (last four).}
    	\label{fig:datasets}
\end{figure*}

\section{Datasets}

In Fig~\ref{fig:datasets} we would like to demonstrate all the objects we employed for our experiments. Thereby, the upper row illustrates all objects of the \textit{unambiguous} dataset, taken from `LineMOD'~\cite{Hinterstoisser2011}. These objects do not exhibit any views which might induce ambiguities. On the contrary, the lower row depicts all objects of the 
\textit{ambiguous} dataset. While the first two objects also belong to the `LineMOD' dataset, the last four accompany the T-LESS dataset~\cite{Hodan2017}. All these objects can induce ambiguities for certain viewpoints. For instance `obj 04' is a symmetric screw, however, possessing distinct textures on its head. Due to this only the views from the bottom (which do not show the texture) are ambiguous. In contrast, for each viewpoint in `obj 09' and `obj 10', there exists always one identical viewpoint on the other side. Thus, these objects are never ambiguity-free.

\newpage

\section{Robust Ambiguity Detection and Estimation}

\begin{table}[H]
    \centering
        \begin{tabular}{@{}c|c|c|c|c|c|c|c|c|c|c@{}}
            & ape & bvise & cam & can & cat & driller & duck & holep & iron & phone  \\
            \midrule
            Ambiguity Detection Accuracy [\%] & 99.9 &	99.9 & 99.9 & 99.8 & 99.7 & 99.7 & 99.6 & 99.1 & 100 & 100  \\
        \end{tabular}\label{tab:lm_unambig}
\end{table}

\begin{table}[H]
    \centering
        \begin{tabular}{@{}c|c|c|c|c|c|c@{}}
            & `eggb' & `glue' & `obj 04' & `obj 05' & `obj 09' & `obj 10' \\
            \midrule
            Ambiguity Detection Accuracy [\%] & 50.4 & 86.6  & 90.3 & 94.3 & 100  & 71.2 \\
            Mean Symmetry Axis Deviation [\degree] & 8.23 & 28.0 & 21.3 & 22.1 & 38.9 & 21.3 \\
            Meanshift Bin Size & $\frac{\pi}{4}$ & $\frac{\pi}{4}$ & $\frac{\pi}{2}$ & $\frac{\pi}{5}$ & $\frac{\pi}{8}$ & $\frac{\pi}{10}$ \\
        \end{tabular}
    \centering
    \caption{Top: Individual ambiguity detection accuracies for the \textit{unambiguous} dataset. Bottom: Individual ambiguity detection accuracies and mean axis deviations for the \textit{ambiguous} dataset.}\label{tab:tless_axis_estimation}
\end{table}

\begin{figure}[H]
	\centering
	
	\includegraphics[width=.24\linewidth]{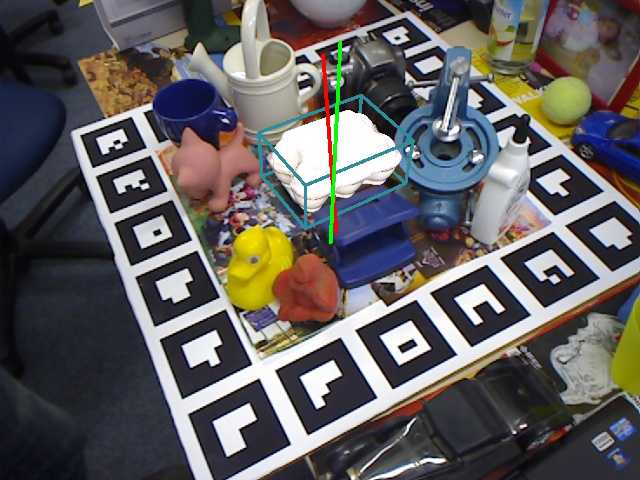}
	\includegraphics[width=.24\linewidth]{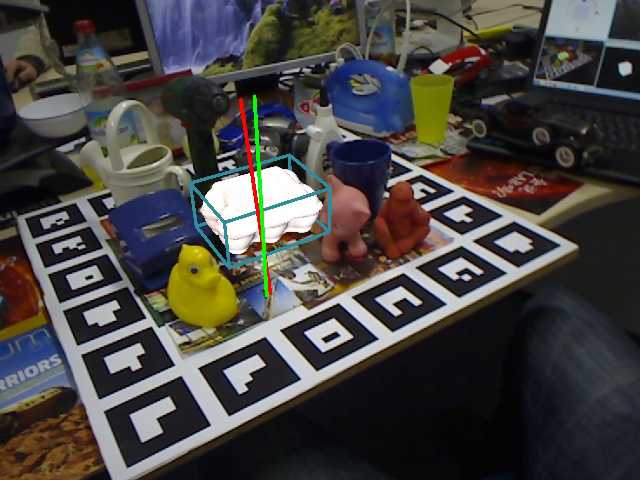}
	\hspace{1mm}
	\includegraphics[width=.24\linewidth]{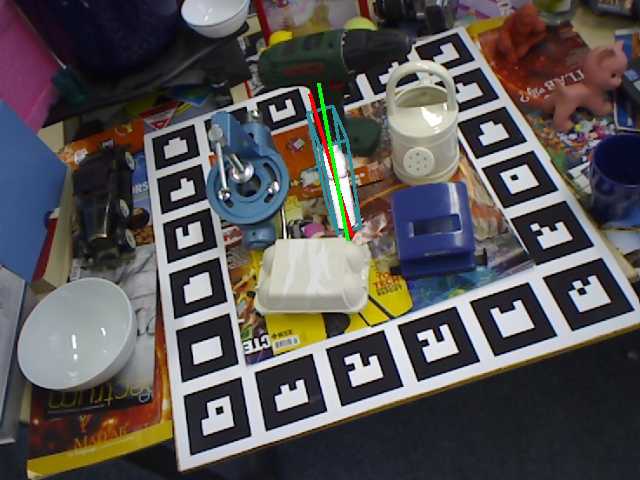}
	\includegraphics[width=.24\linewidth]{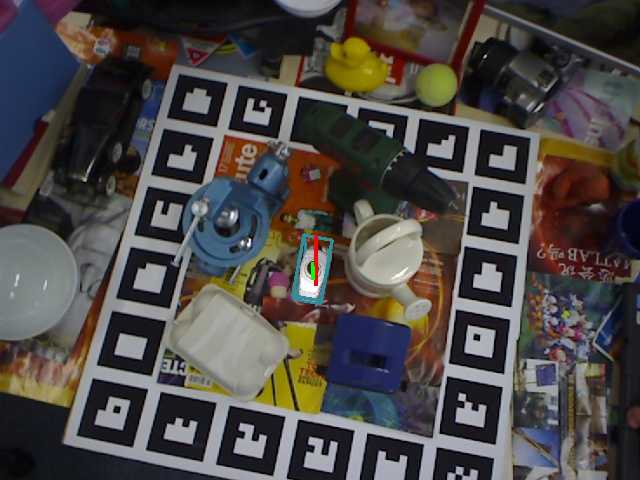}\\
	\includegraphics[width=.24\linewidth]{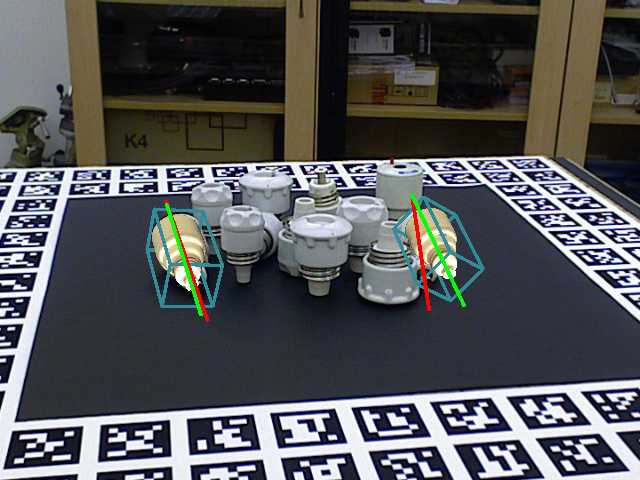}
	\includegraphics[width=.24\linewidth]{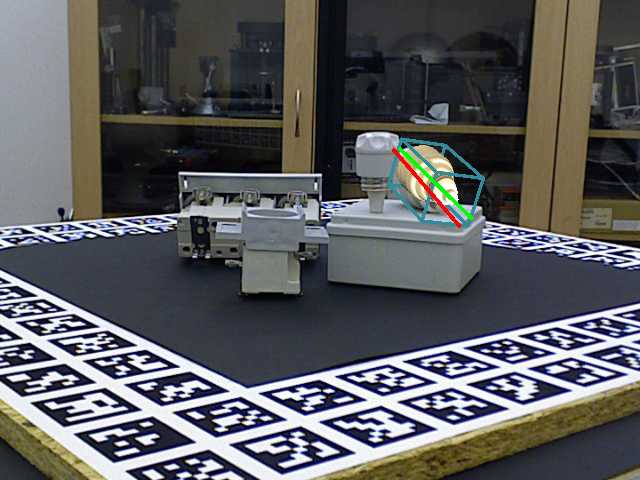}
	\hspace{1mm}
	\includegraphics[width=.24\linewidth]{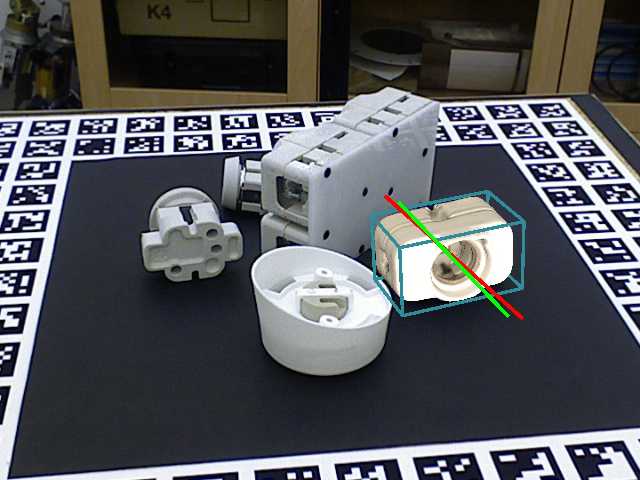}
	\includegraphics[width=.24\linewidth]{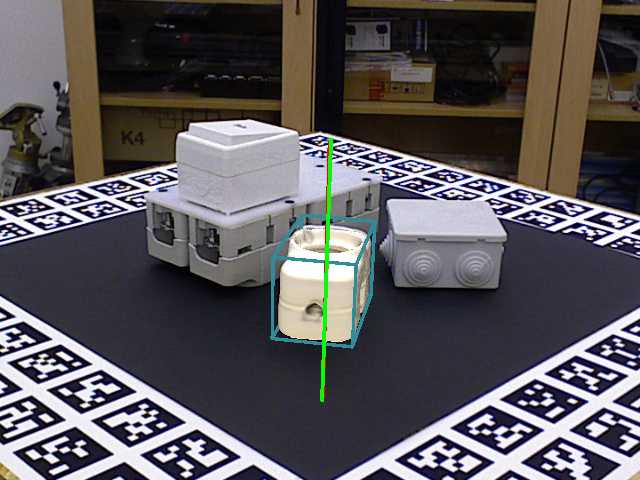}
	\centering
	\includegraphics[width=.24\linewidth]{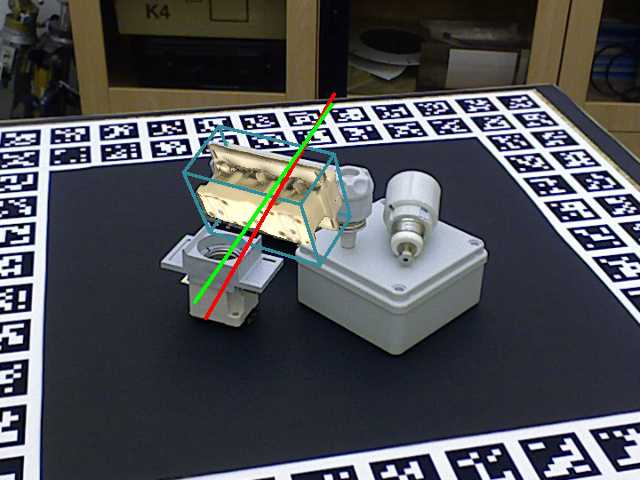}
	\includegraphics[width=.24\linewidth]{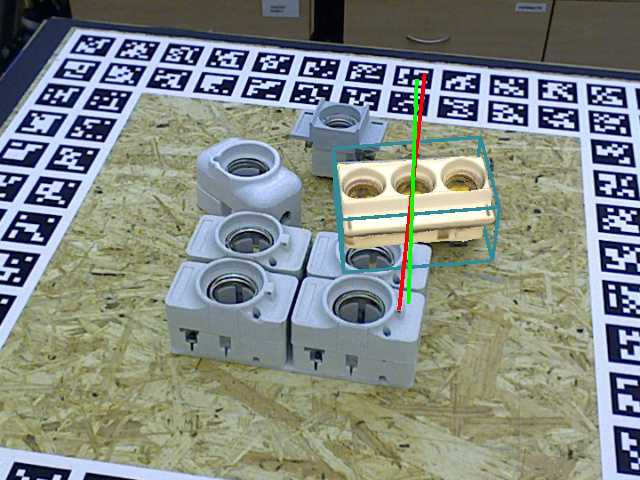}
	\hspace{1mm}
	\includegraphics[width=.24\linewidth]{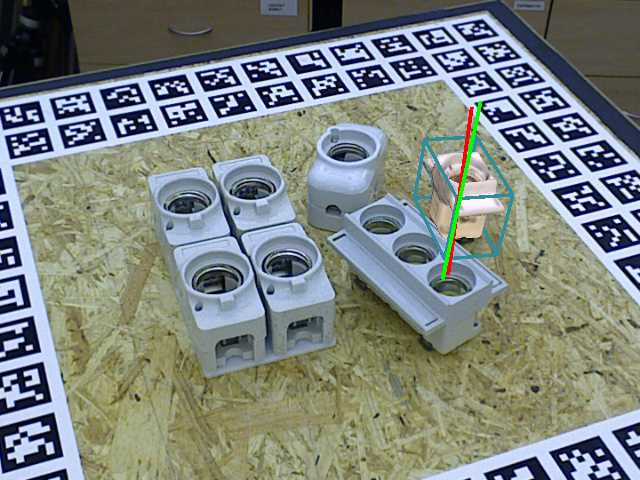}
	\includegraphics[width=.24\linewidth]{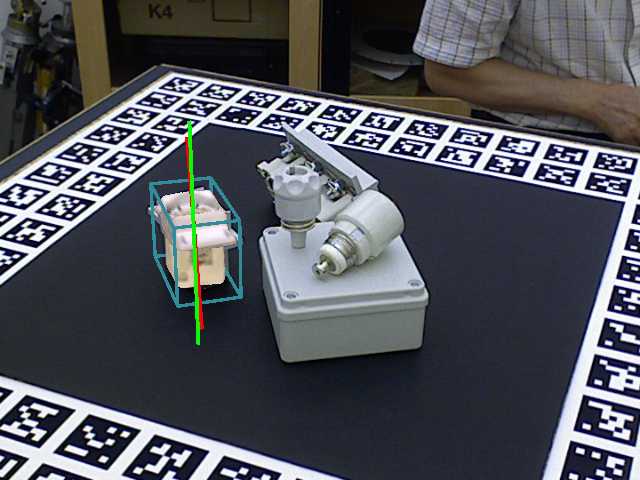}
	\\
	\caption{Qualitative samples for ambiguity detection and ambiguity axis estimation. The green line illustrates the computed axis and the red axis depicts the ground truth axis.}
	\label{fig:am_axis}
\end{figure}

Tab.\ref{tab:tless_axis_estimation} shows our detailed ambiguity detection results for the unambiguous (top) and ambiguous (bottom) objects, respectively. In addition, we also report our individual results for the ambiguity axis estimation. We compute the mean deviation from the labeled ground truth. As a threshold for $\sigma_1$ we empirically find $0.8$ to offer good accuracy. Fig.~\ref{fig:am_axis} demonstrates more qualitative results for ambiguity detection and the computation of the corresponding ambiguity axis.

\newpage

\section{2D Object Detection and 6D Pose Estimation}

In this section, we present our detailed results for 6D pose estimation and 2D detection. As in the paper, for the \textit{unambiguous} dataset we present our numbers with $M=5$ and for the \textit{ambiguous} dataset we set $M=30$.

\subsection{Unambiguous Object Detection and Pose Estimation}

We present an ablation study for different numbers of hypotheses $\mathbf{M}$ in Tab. \ref{tab:ablation}. We obtained our best results employing $\mathbf{M} = 5$ hypotheses. Below we show one qualitative sample for each object. In addition, on the right we also visualize the corresponding Bingham Distributions for visual validation. Lastly, we depict some qualitative results on the `LineMOD Occlusion' dataset.

\begin{table}[H]
	\begin{center}
	
	\begin{tabular}{@{}c||c|c|c|c||c@{}}
		& Rot. [\degree] & Trans. [mm] & VSS [\%]& ADD [\%] & F1 \\ \midrule
		$\mathbf{M} = 1$ & 17.9 & 45.6 & 76.8 & 31.2 & 91.6 \\
	    $\mathbf{M} = 2$ & 18.9 & 44.3 & 76.3 & 32.8 & 92.1 \\
		$\mathbf{M} = 5$ & \textbf{17.4} & \textbf{39.5} & \textbf{78.2} & \textbf{35.3} & \textbf{93.4} \\
		$\mathbf{M} = 10$ & 19.2 & 45.6 & 77.2 & 31.3 & 90.6 \\
		$\mathbf{M} = 20$ & 18.7 & 44.6 & 77.4 & 33.8 & 92.7 \\
		$\mathbf{M} = 30$ & 19.2 & 44.9 & 77.3 & 32.7 & 91.0 \\
		$\mathbf{M} = 40$ & 22.5 & 42.6 & 77.4 & 35.7 & 91.0 \\
		\midrule
	\end{tabular}
     
	\end{center}
	\caption{Ablation study on the impact of different number of hypotheses.}\label{tab:ablation}
\end{table}

\begin{figure}[H]
	\centering
	\includegraphics[width=.32\linewidth]{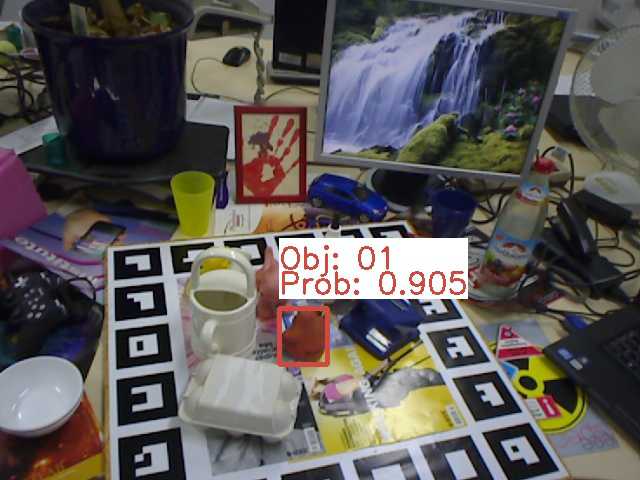}
	\includegraphics[width=.32\linewidth]{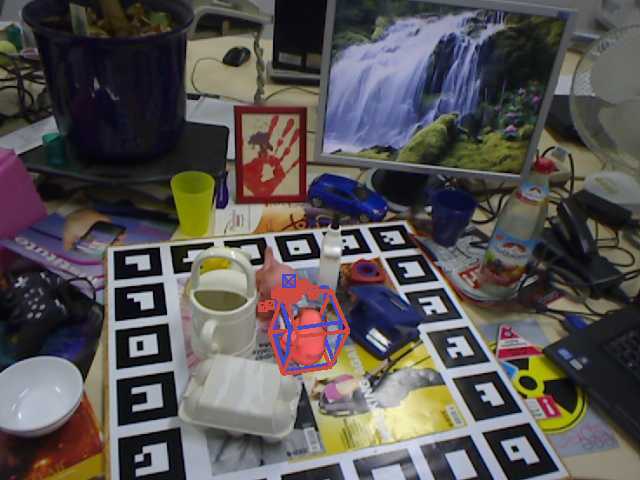}
	\includegraphics[width=.32\linewidth]{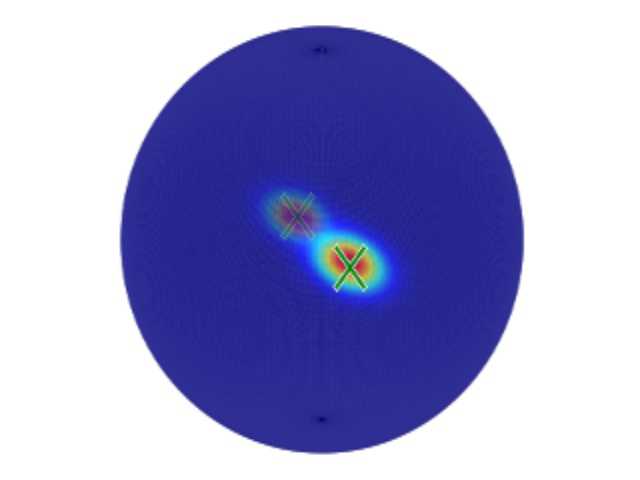} \\
	\includegraphics[width=.32\linewidth]{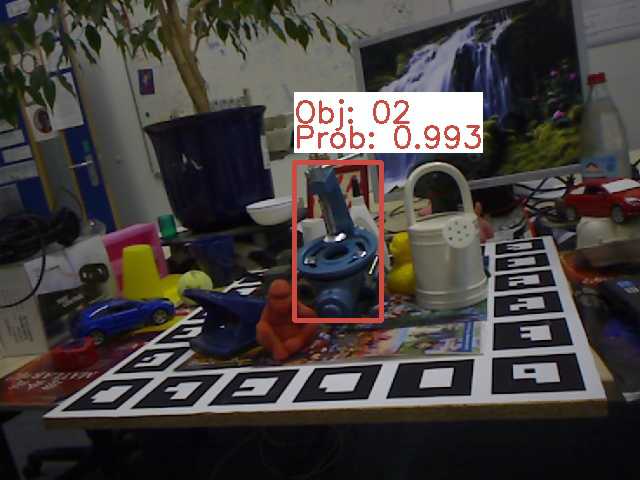}
	\includegraphics[width=.32\linewidth]{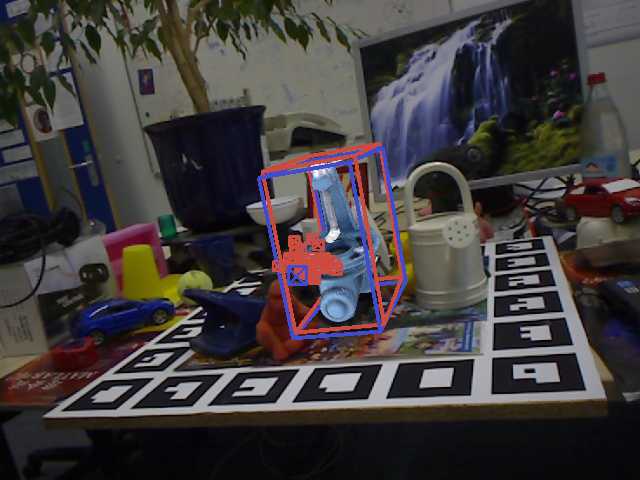}
	\includegraphics[width=.32\linewidth]{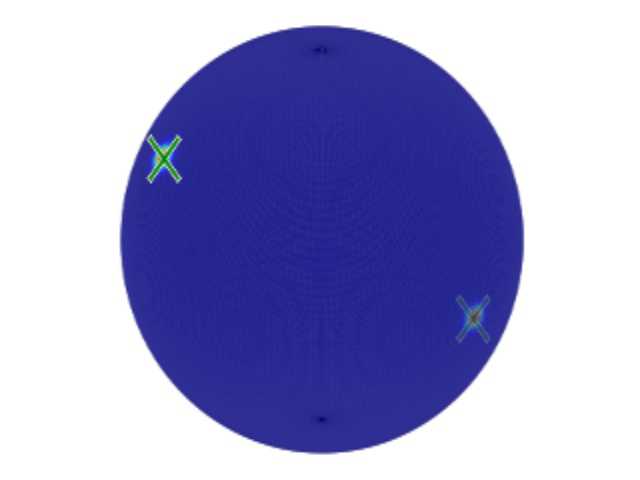}
\end{figure}

\begin{figure}[H]
	\centering
	\includegraphics[width=.32\linewidth]{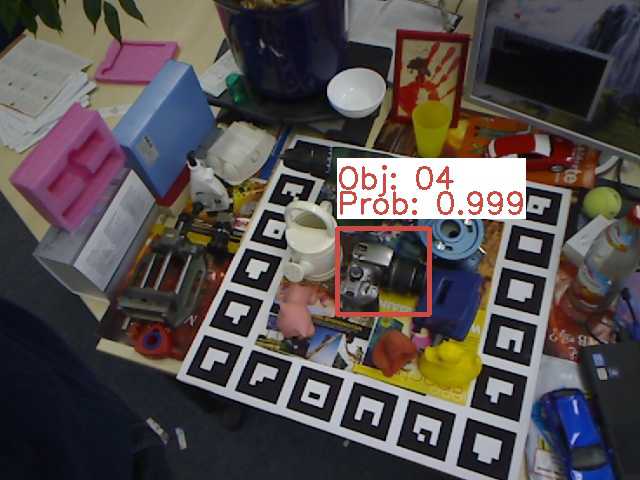}
	\includegraphics[width=.32\linewidth]{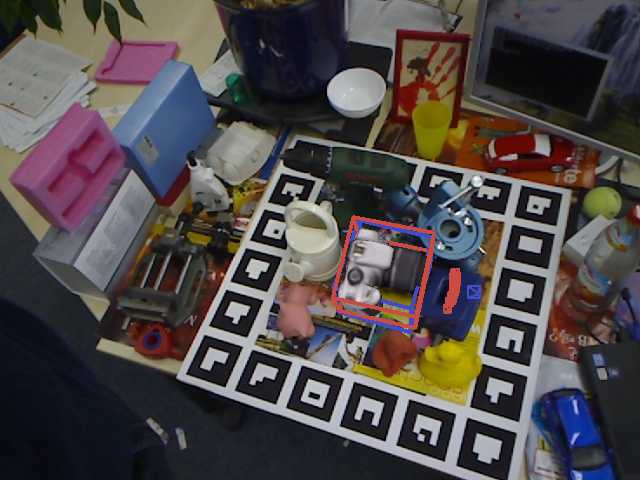}
	\includegraphics[width=.32\linewidth]{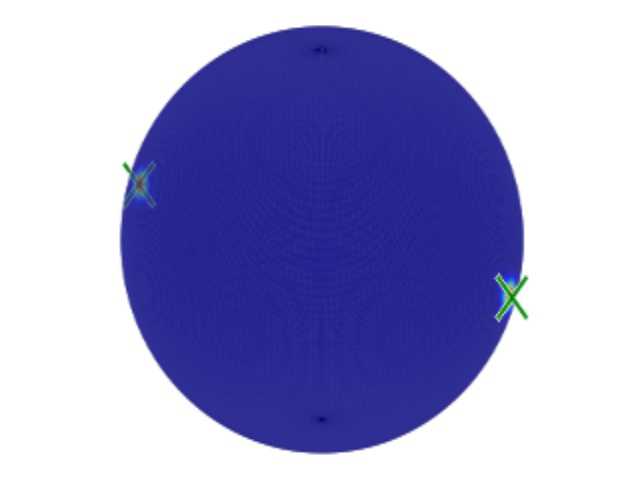} \\
	\includegraphics[width=.32\linewidth]{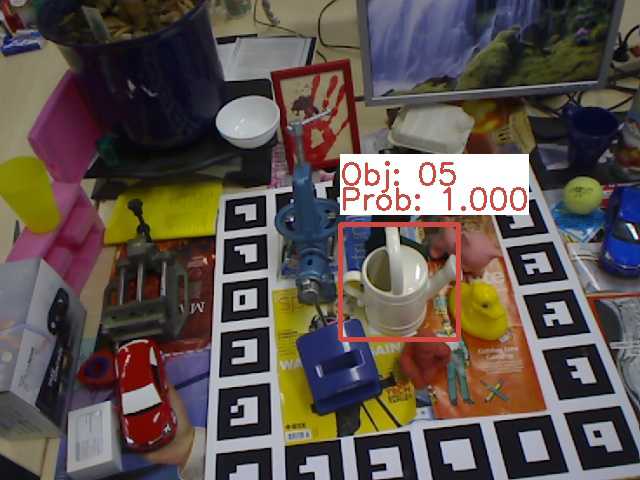}
	\includegraphics[width=.32\linewidth]{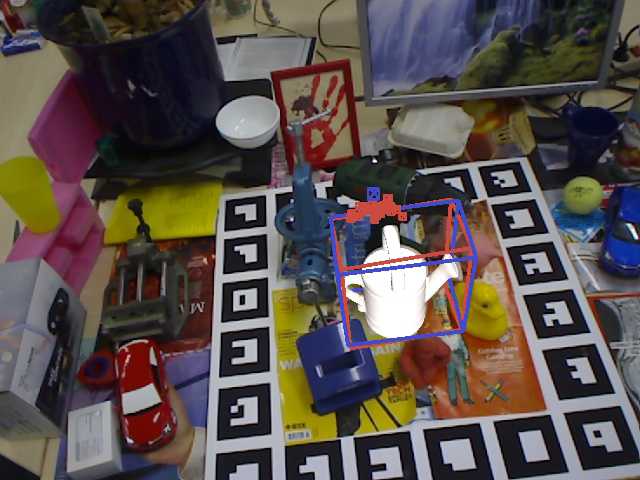}
	\includegraphics[width=.32\linewidth]{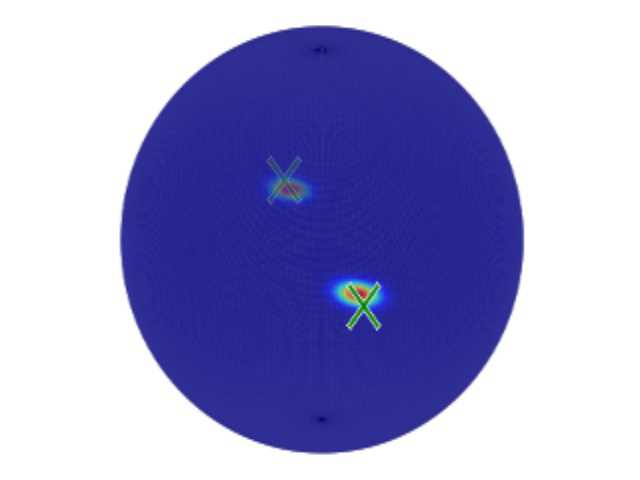} \\
	\includegraphics[width=.32\linewidth]{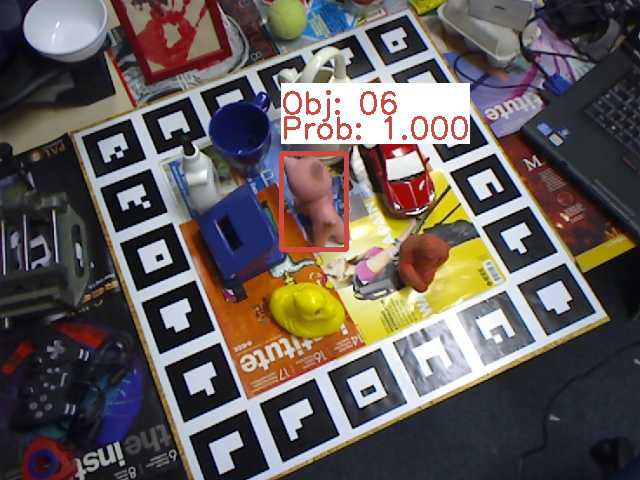}
	\includegraphics[width=.32\linewidth]{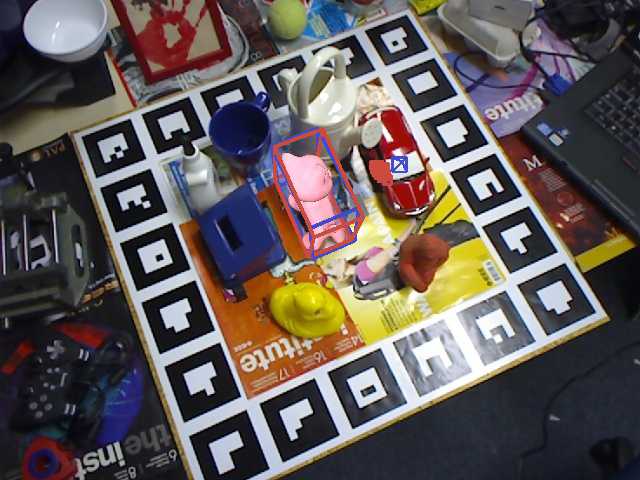}
	\includegraphics[width=.32\linewidth]{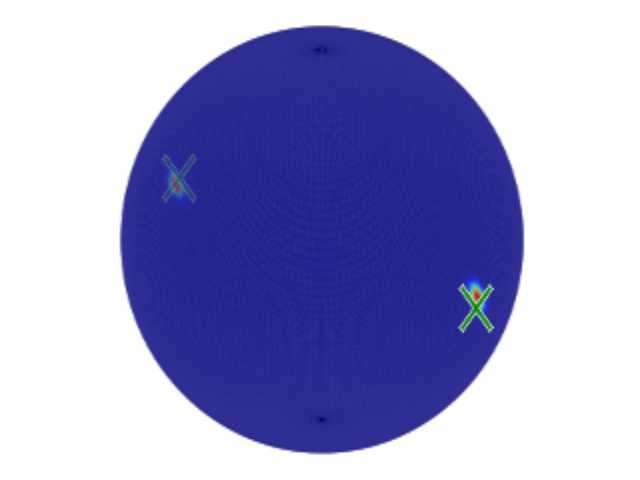} \\
	\includegraphics[width=.32\linewidth]{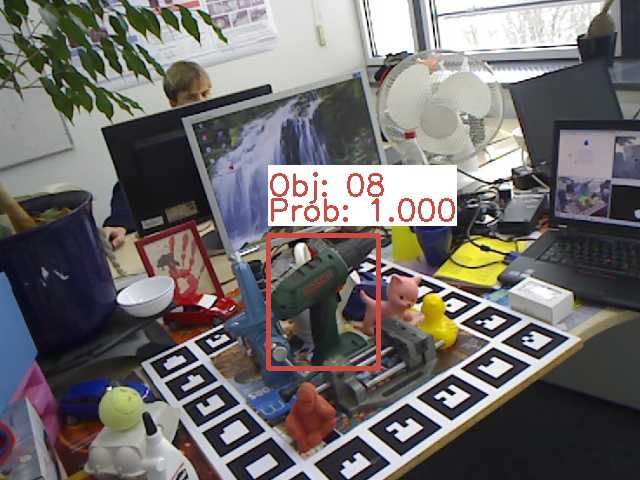}
	\includegraphics[width=.32\linewidth]{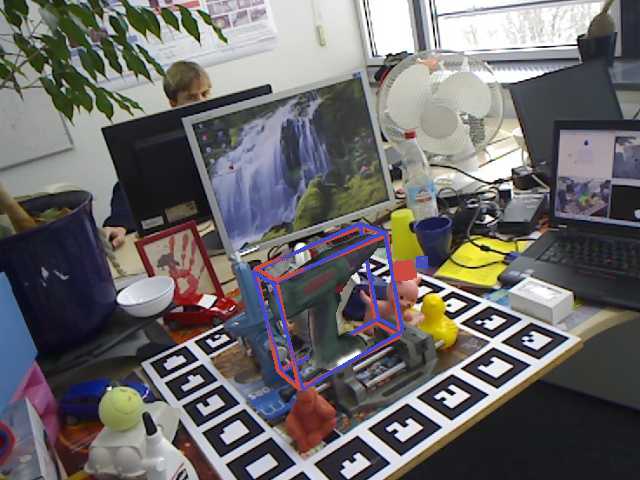}
	\includegraphics[width=.32\linewidth]{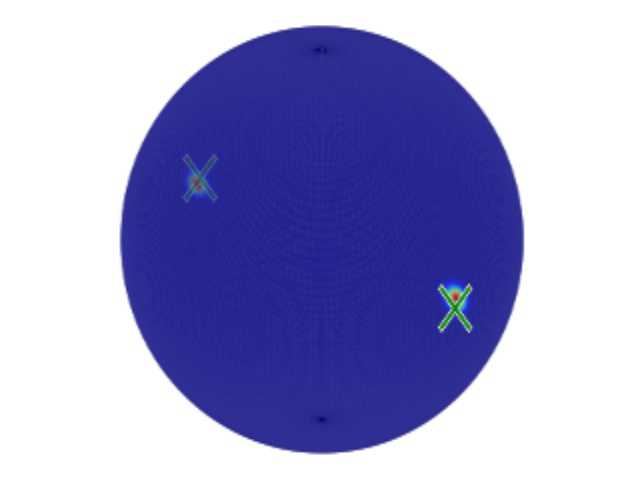} \\
	\includegraphics[width=.32\linewidth]{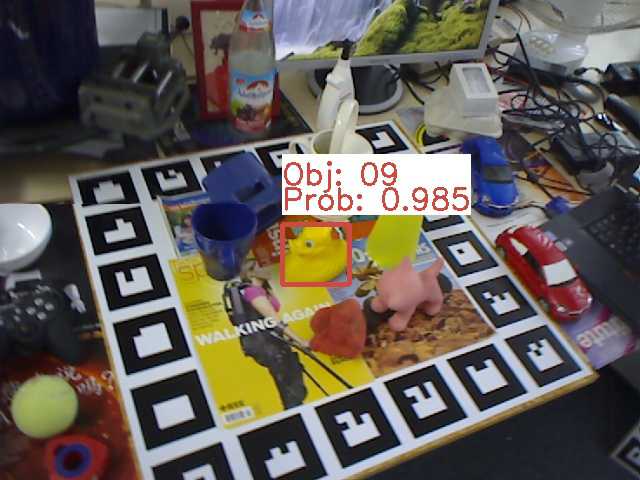}
	\includegraphics[width=.32\linewidth]{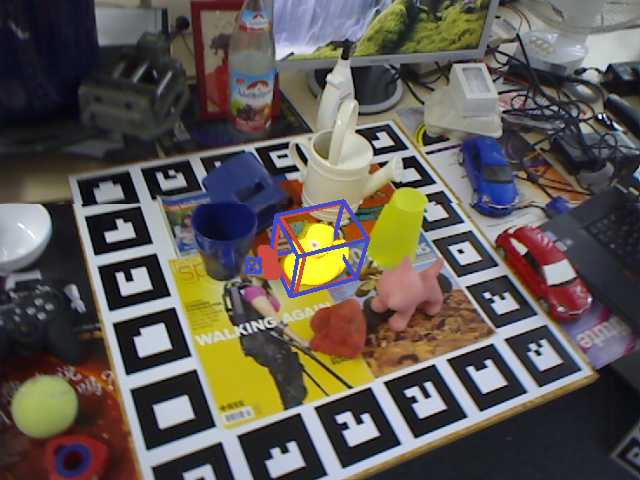}
	\includegraphics[width=.32\linewidth]{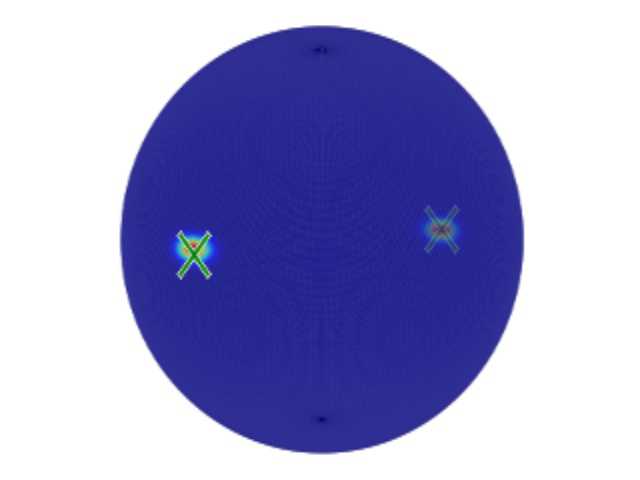} 
\end{figure}

\begin{figure}[H]
	\centering
	\includegraphics[width=.32\linewidth]{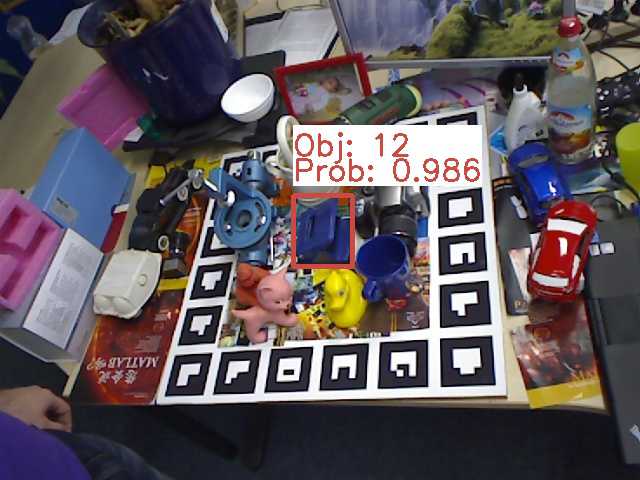}
	\includegraphics[width=.32\linewidth]{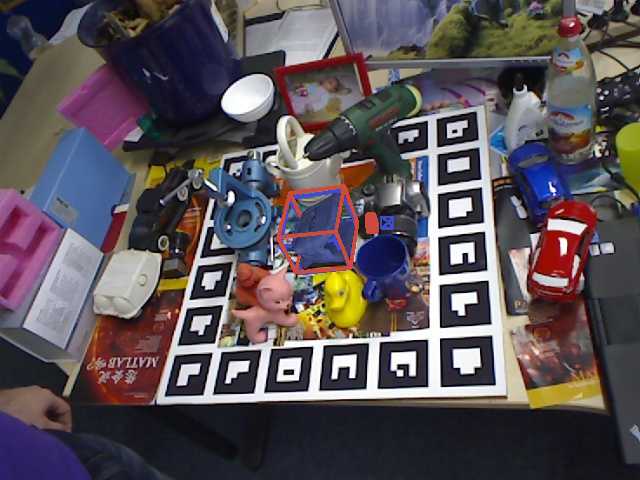}
	\includegraphics[width=.32\linewidth]{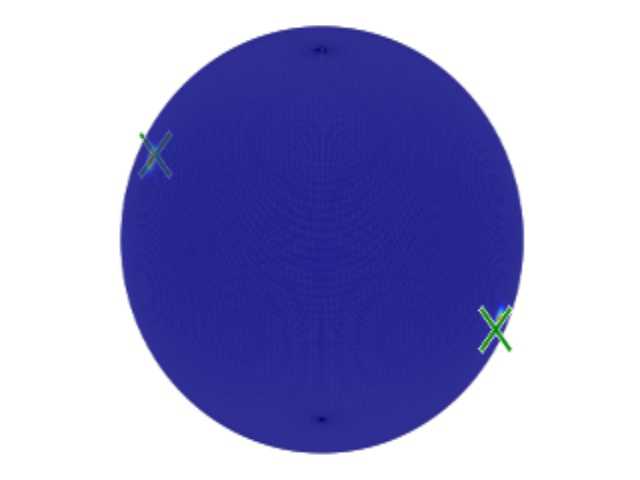} \\
	\includegraphics[width=.32\linewidth]{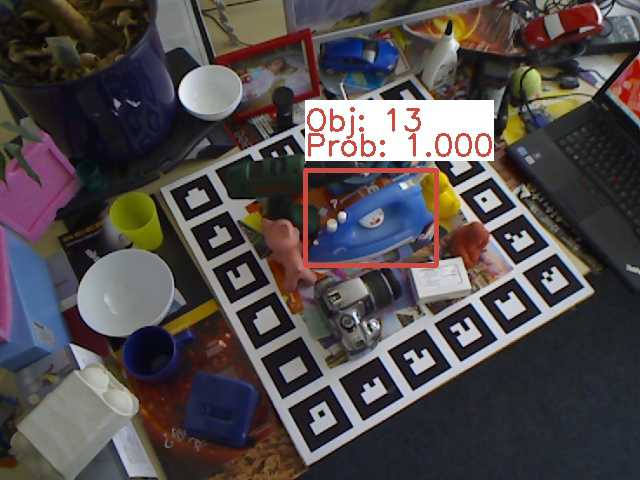}
	\includegraphics[width=.32\linewidth]{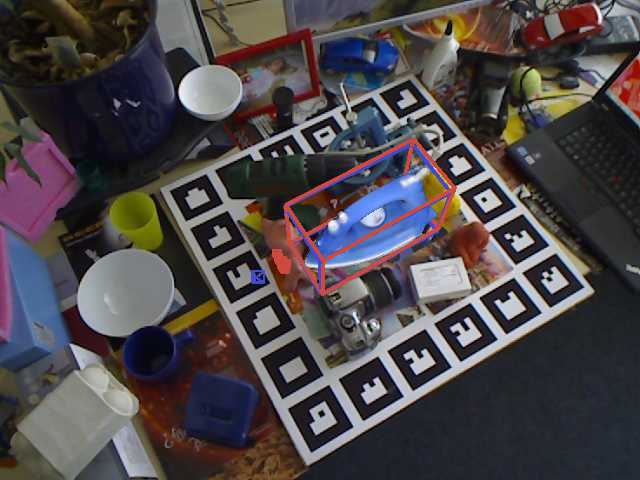}
	\includegraphics[width=.32\linewidth]{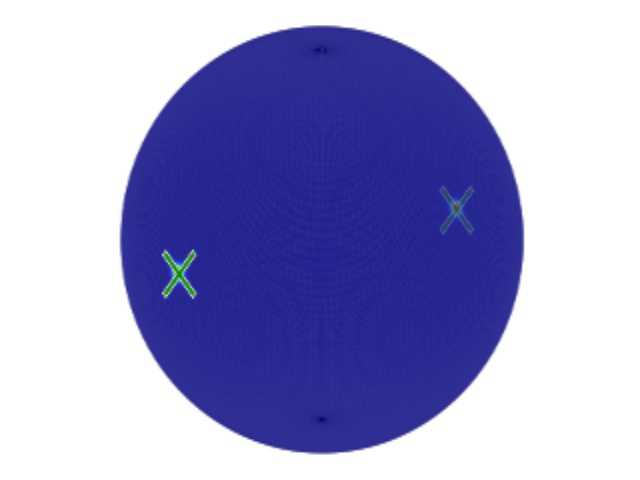} \\ \vspace{-3.5mm}
	\subfloat[2D Detections]{\includegraphics[width=.32\linewidth]{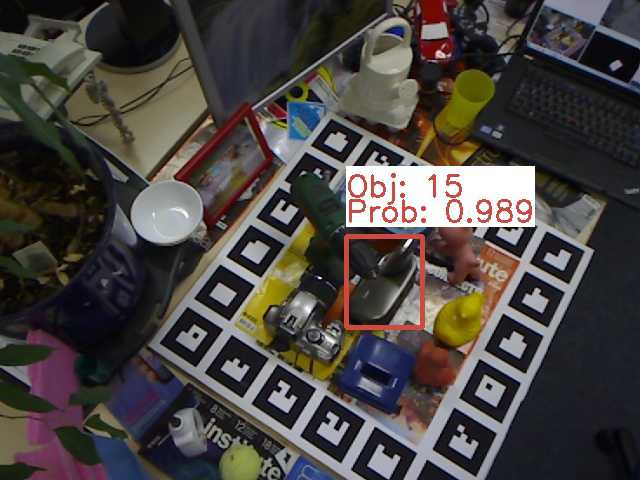}}\hspace{0.066mm}
	\subfloat[6D Pose and Associated Hypotheses]{\includegraphics[width=.32\linewidth]{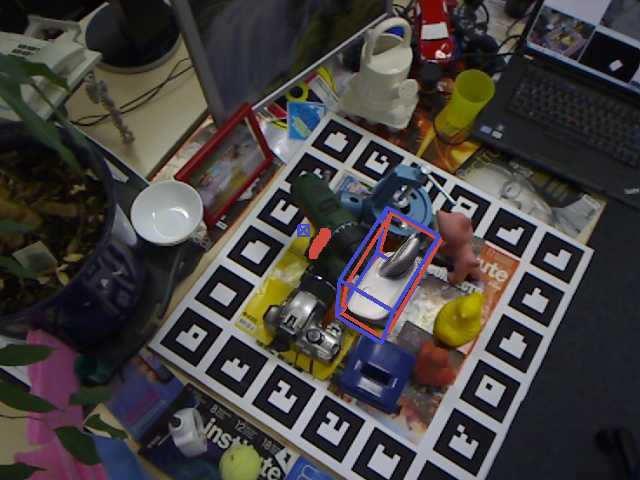}} \hspace{0.066mm}
	\subfloat[Bingham Distributions]{\includegraphics[width=.32\linewidth]{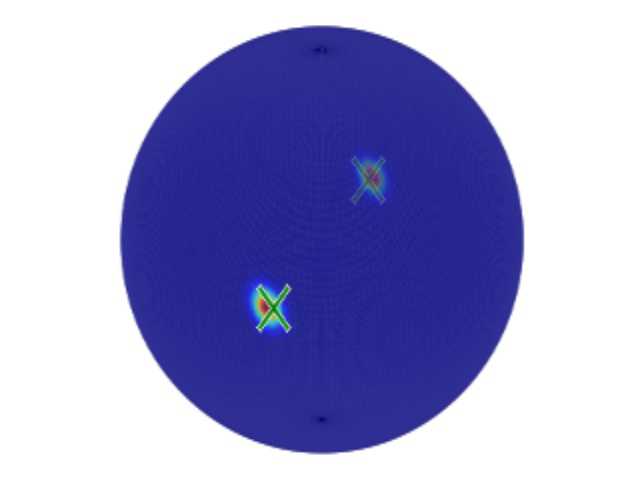}}\\
	\caption{Qualitative results for the unambiguous objects.}
	\label{fig:train_samples}
\end{figure}

\begin{figure}[H]
	\centering
	\includegraphics[width=.32\linewidth]{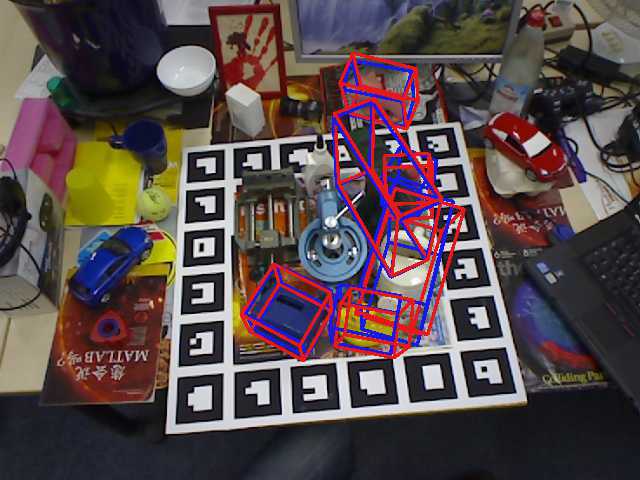}
	\includegraphics[width=.32\linewidth]{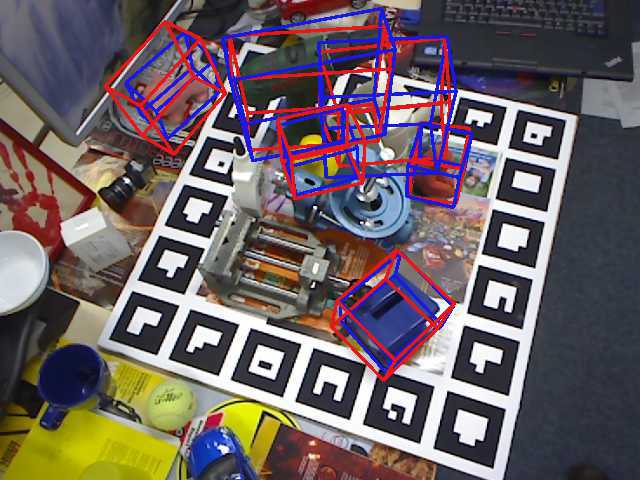}
	\includegraphics[width=.32\linewidth]{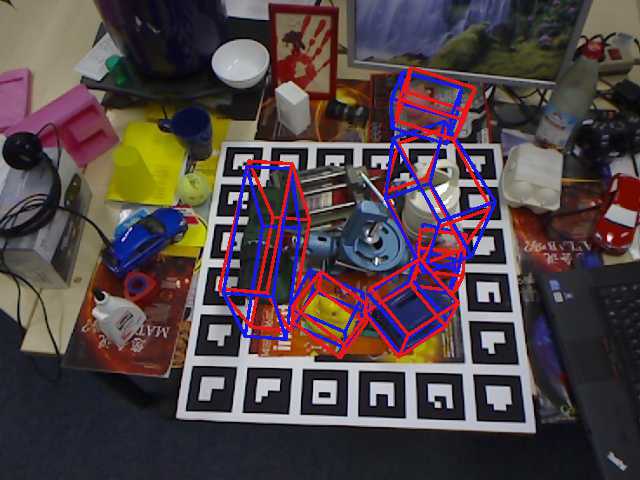} \\
	\includegraphics[width=.32\linewidth]{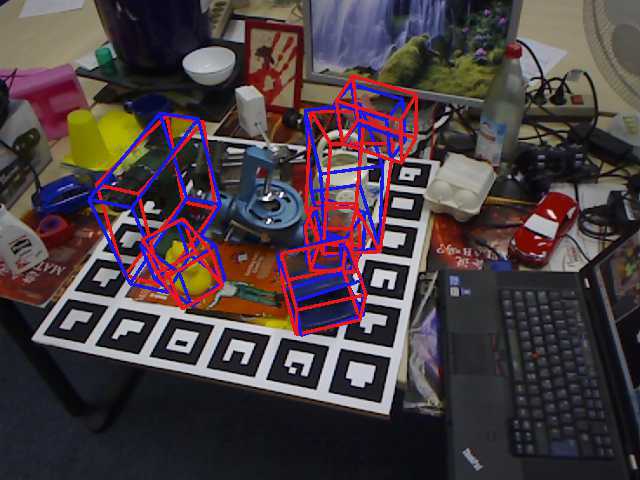}
	\includegraphics[width=.32\linewidth]{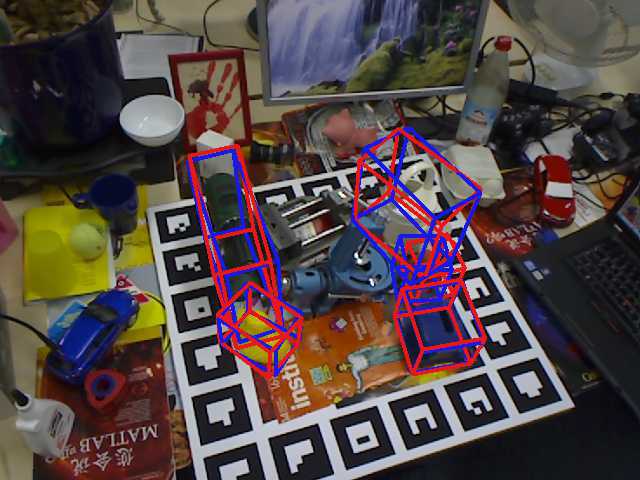}
	\includegraphics[width=.32\linewidth]{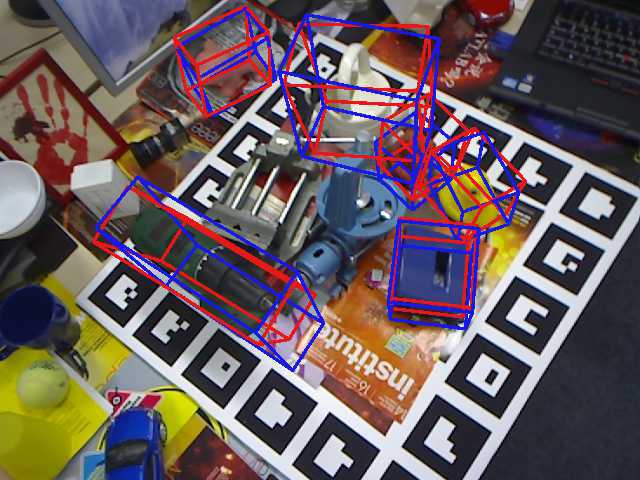} \\
	\includegraphics[width=.32\linewidth]{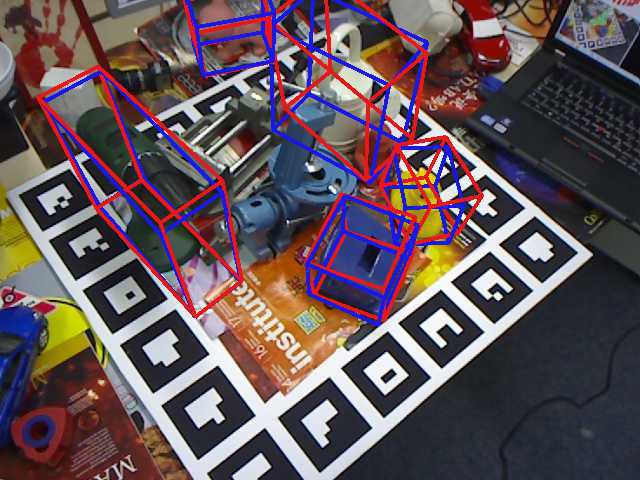}
	\includegraphics[width=.32\linewidth]{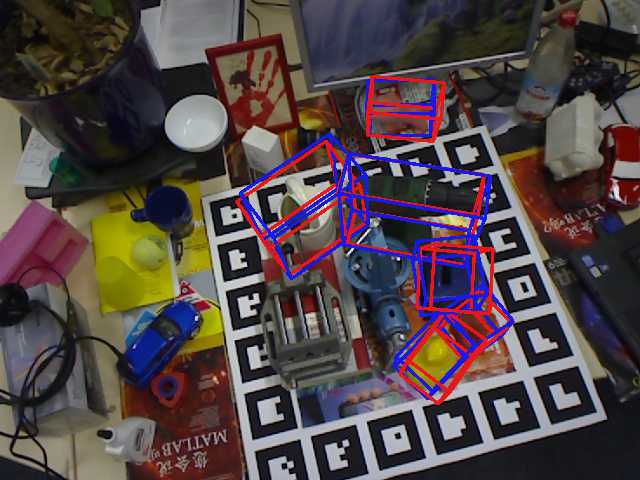}
	\includegraphics[width=.32\linewidth]{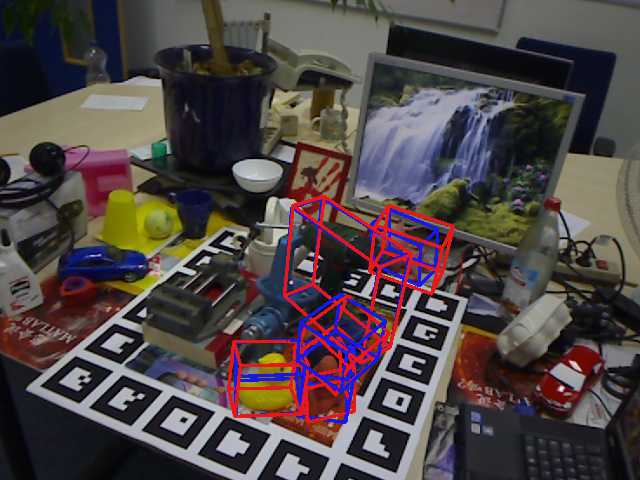} \\
	\includegraphics[width=.32\linewidth]{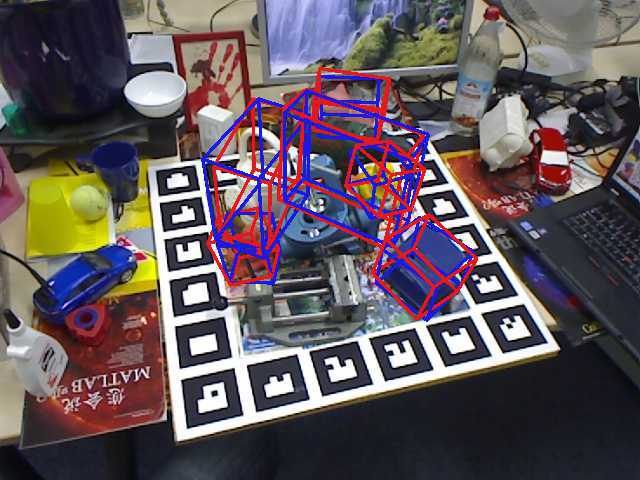}
	\includegraphics[width=.32\linewidth]{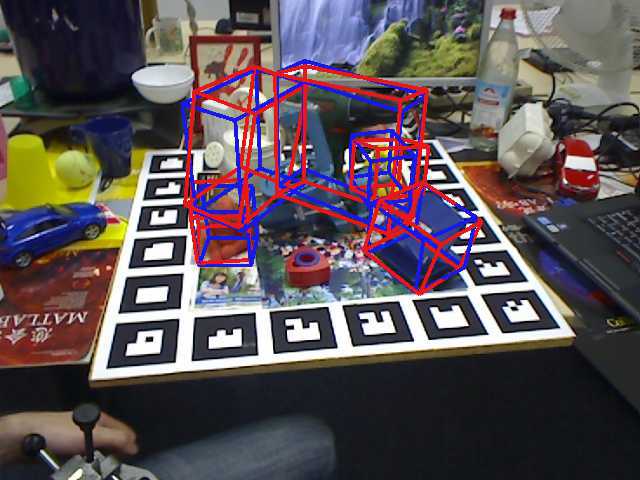}
	\includegraphics[width=.32\linewidth]{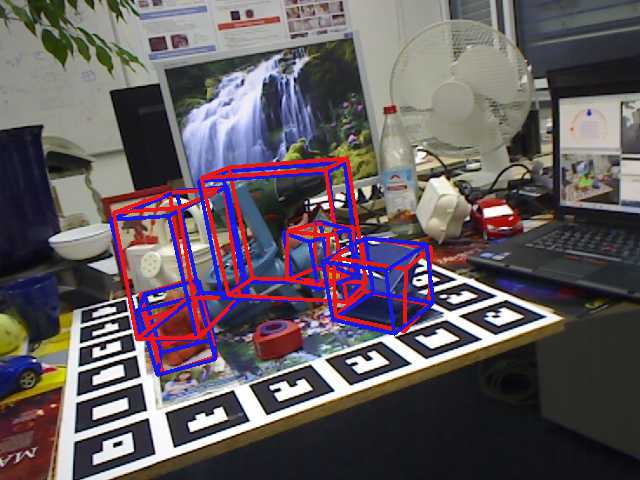} \\
	\includegraphics[width=.32\linewidth]{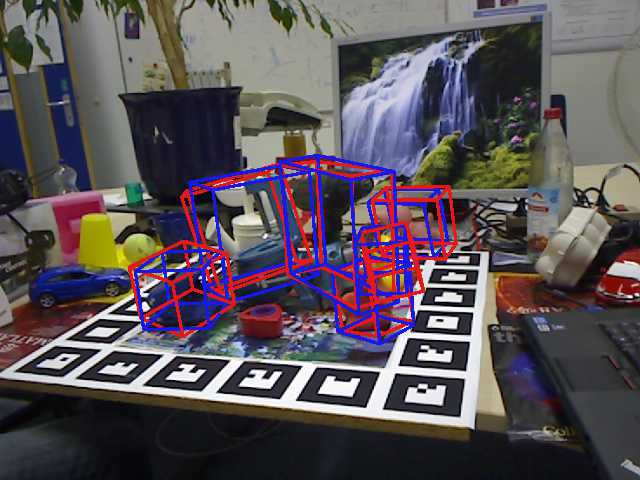}
	\includegraphics[width=.32\linewidth]{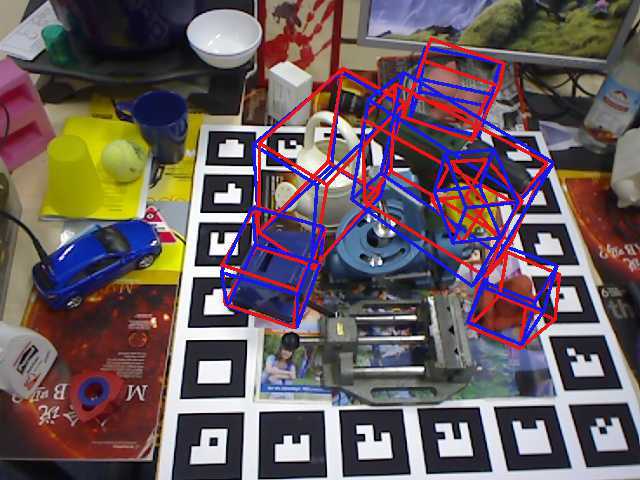}
	\includegraphics[width=.32\linewidth]{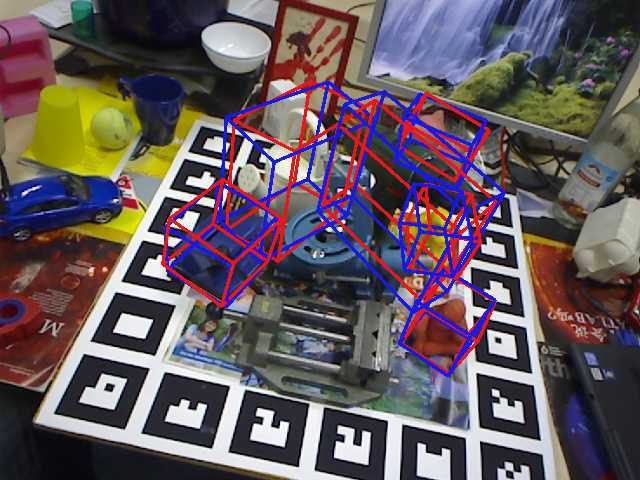} 
	\caption{Qualitative results for `LineMOD Occlusion'}
	\label{fig:lm_occlusion}
\end{figure}

\subsection{Ambiguous Object Detection and Pose Estimation}

\begin{table}[H]
	\begin{center}
    		\begin{tabular}{@{}c|c|c|c|c|c|c|c|c|c|c|c@{}}
    			Object & 'eggb' & 'glue' & \multicolumn{2}{c|}{obj\_04} & \multicolumn{3}{c|}{obj\_05} & \multicolumn{2}{c|}{obj\_09} & \multicolumn{2}{c}{obj\_10}\\
    			Scene & -- & -- & 5 & 9 & 2 & 3 & 9 & 5 & 11 & 5 & 11 \\ \hline\hline
    			ADI [\%] & 55.7 & 54.6 & 17.1 & 22.1 & 87.6 & 62.1 & 84.4 & 65.5 & 74.2 & 62.0 & 53.8 \\
    			VSS [\%] & 83.1 & 74.6 & 66.0 & 75.5 & 89.2 & 85.9 & 87.6 & 84.4 & 84.4 & 83.5 & 80.4 \\
    		\end{tabular}
		\caption{Detailed evaluation scores the \textit{ambiguous} dataset.}
		\label{tab:tless_individual_pose}
	\end{center}
\end{table}

Since comparing against the ground truth is not suitable in a multiple hypothesis scenario, only metrics that do not rely on this value are apt for this case. We thus chose the Visual Surface Similarity \cite{Kehl2017, Manhardt_2018_ECCV} and Average Distance of Indistinguishable points \cite{Hodan2016} as metrics for pose. We always take the detection with the highest confidence. We present our individual scores for the \textit{ambiguous} dataset in Tab~\ref{tab:tless_individual_pose}. Additionally, below we show one qualitative sample for each object.

\begin{figure}[H]
	\centering
	\includegraphics[width=.32\linewidth]{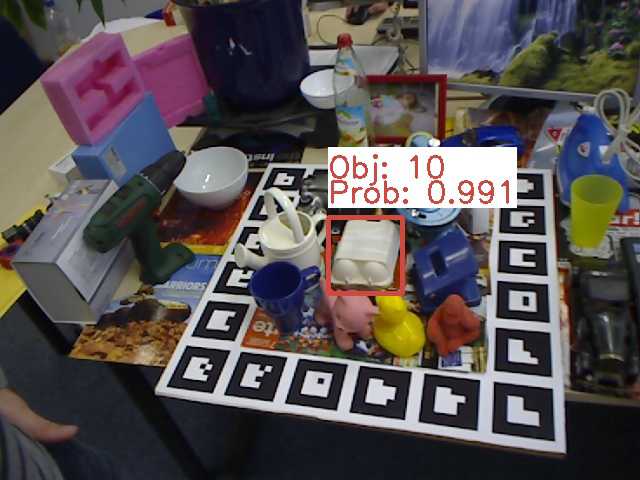}
	\includegraphics[width=.32\linewidth]{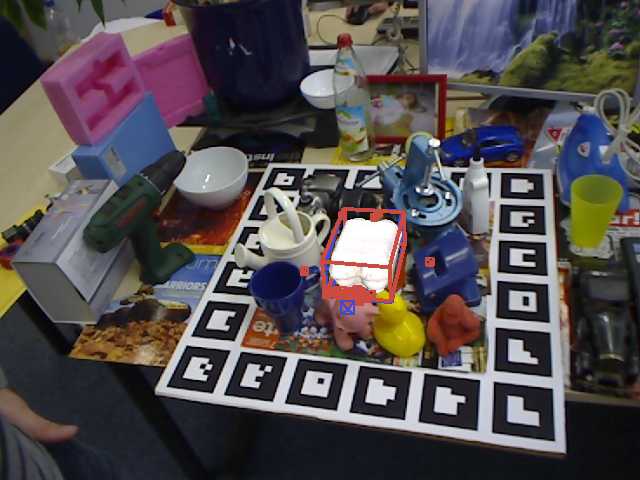}
	\includegraphics[width=.32\linewidth]{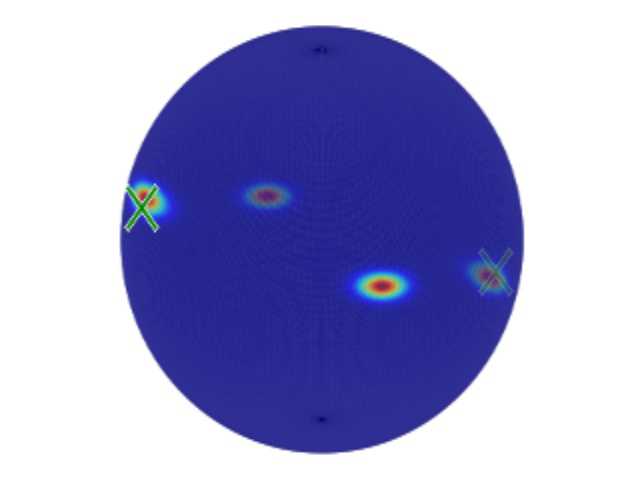} \\
	\includegraphics[width=.32\linewidth]{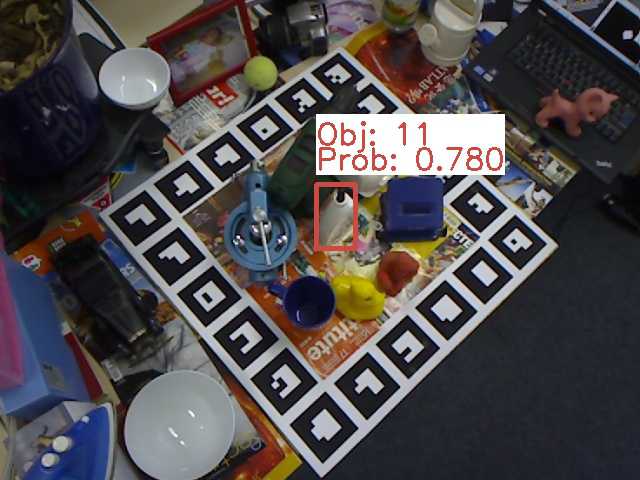}
	\includegraphics[width=.32\linewidth]{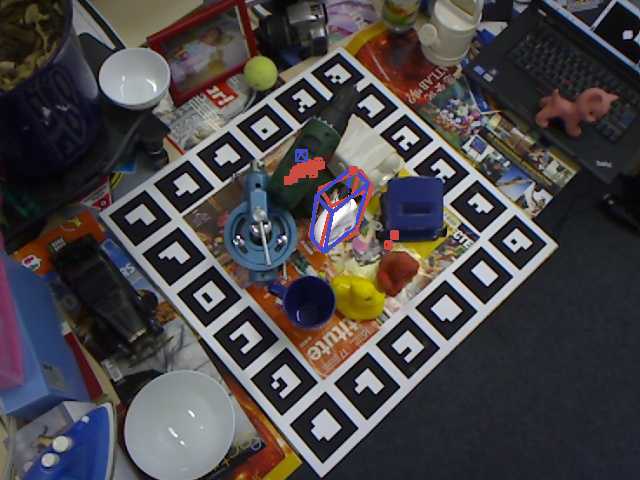}
	\includegraphics[width=.32\linewidth]{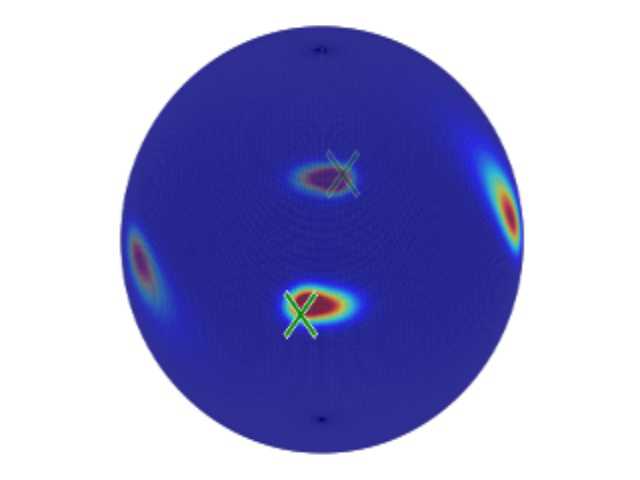} \\
	\includegraphics[width=.32\linewidth]{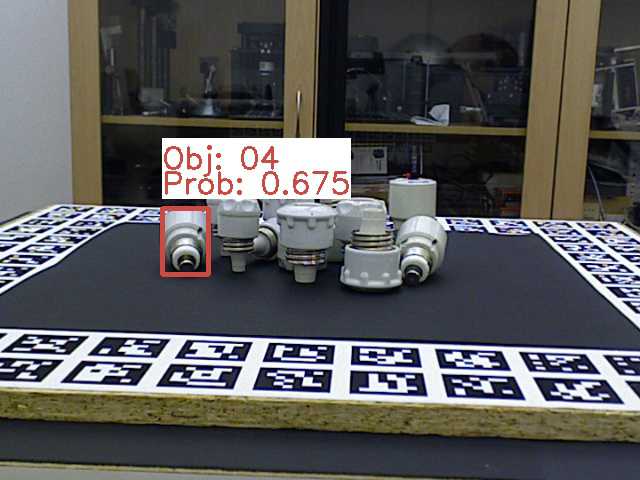}
	\includegraphics[width=.32\linewidth]{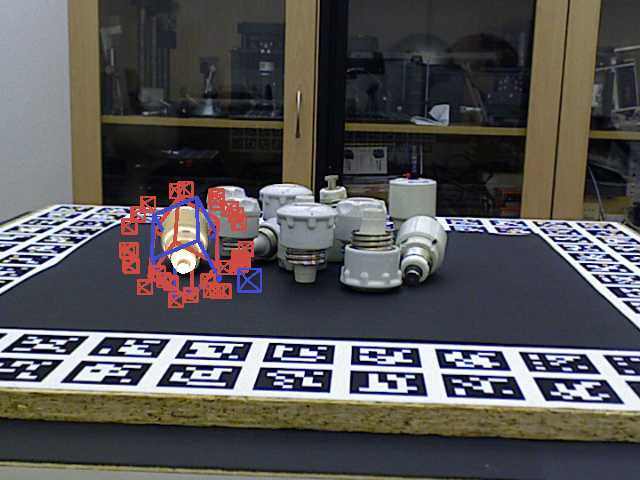}
	\includegraphics[width=.32\linewidth]{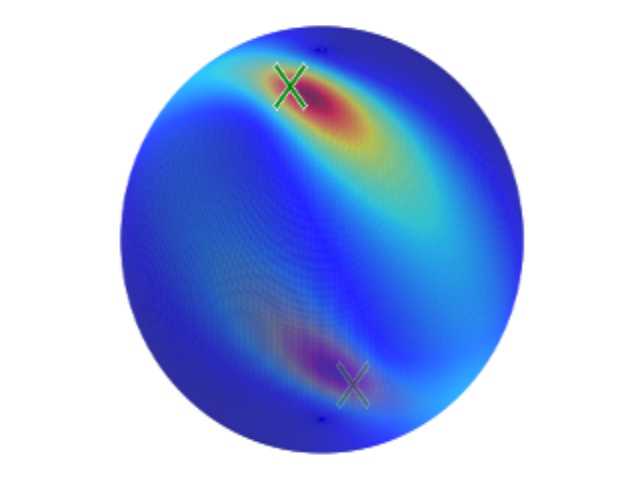} \\
\end{figure}

\begin{figure}[H]
	\centering
	\includegraphics[width=.32\linewidth]{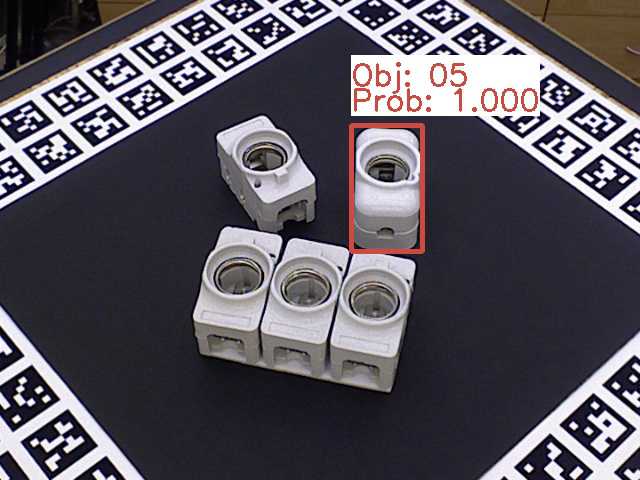}
	\includegraphics[width=.32\linewidth]{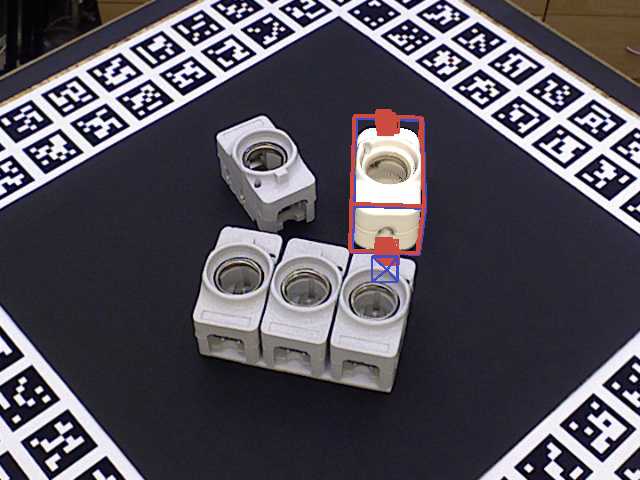}
	\includegraphics[width=.32\linewidth]{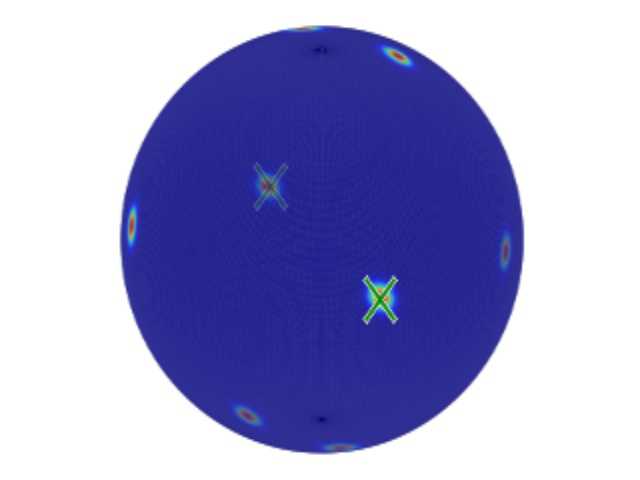} \\
	\includegraphics[width=.32\linewidth]{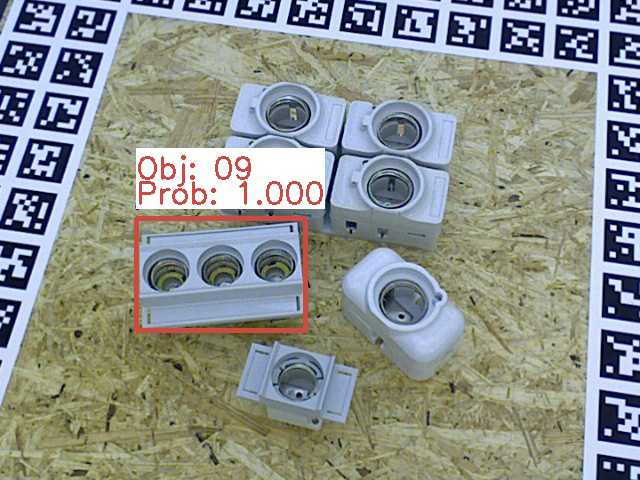}
	\includegraphics[width=.32\linewidth]{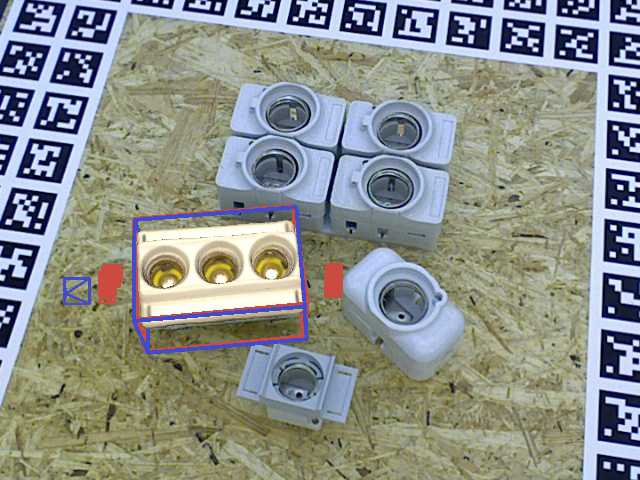}
	\includegraphics[width=.32\linewidth]{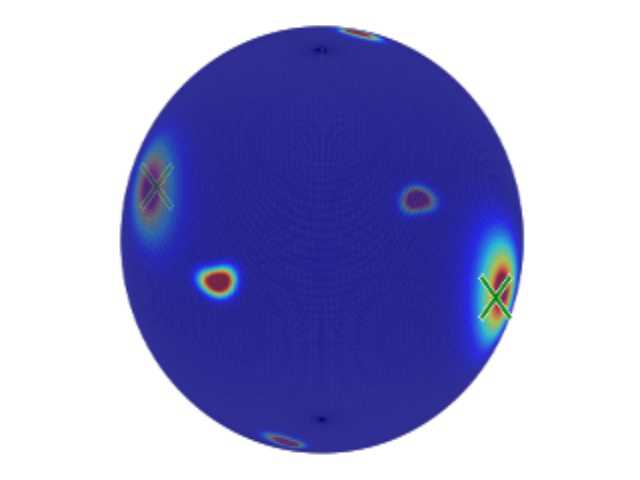} \\  \vspace{-3.5mm}
	\subfloat[Input Image]{\includegraphics[width=.32\linewidth]{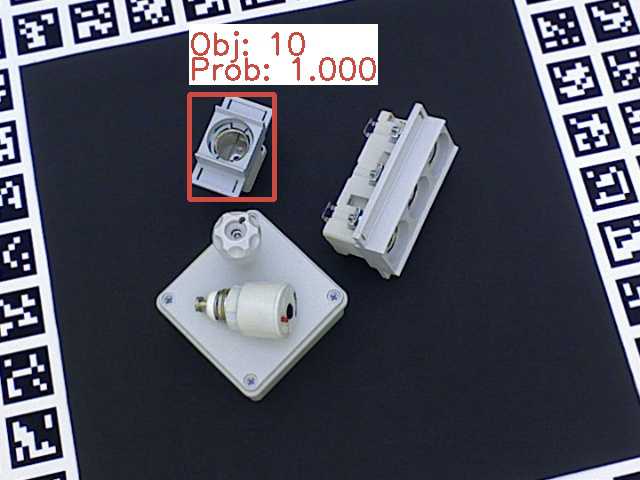}}\hspace{0.066mm}
	\subfloat[2D Detections]{\includegraphics[width=.32\linewidth]{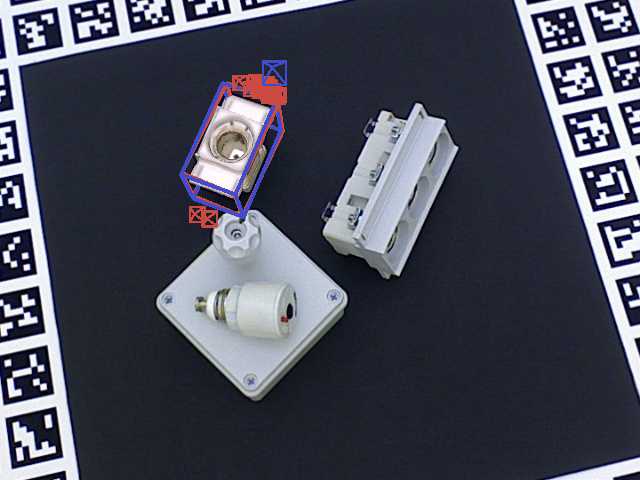}} \hspace{0.066mm}
	\subfloat[6D Pose and Associated Hypotheses]{\includegraphics[width=.32\linewidth]{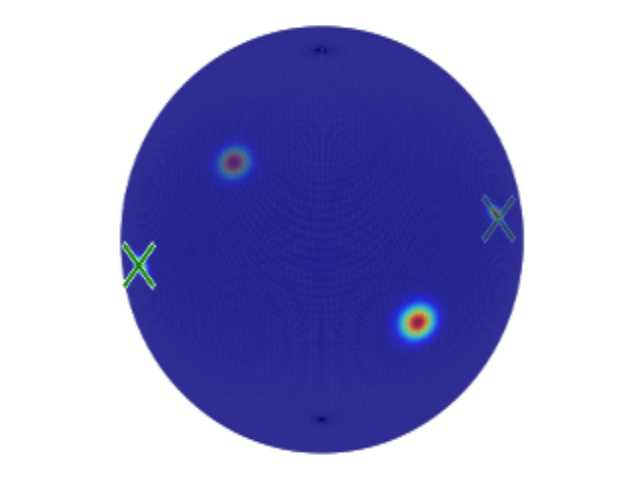}}\\
	\caption{Qualitative results for the ambiguous objects.}
	\label{fig:train_samples}
\end{figure}

\newpage
\section{Employing Multiple Hypothesis as Measurement For Reliability}

We would like to present more qualitative samples that the hypotheses can be employed as measurement for confidence. To this end, for each object of the \textit{unambiguous} dataset we show the poses possessing the lowest and the highest standard deviation in the hypotheses.

\begin{figure}[H]
	\centering
	
	\includegraphics[width=.24\linewidth]{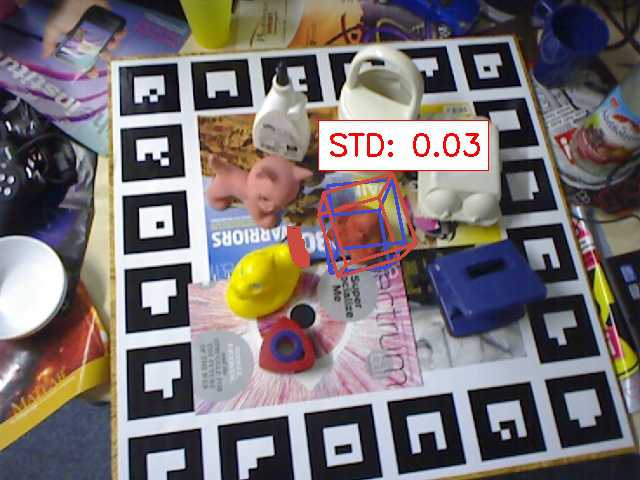}
	\includegraphics[width=.24\linewidth]{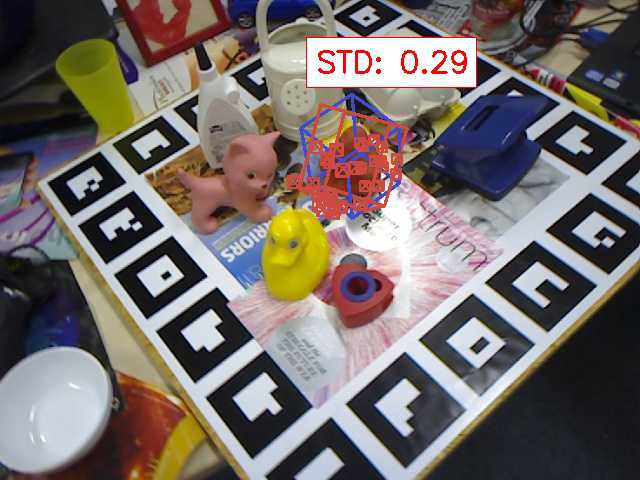}
	\hspace{3mm}
	\includegraphics[width=.24\linewidth]{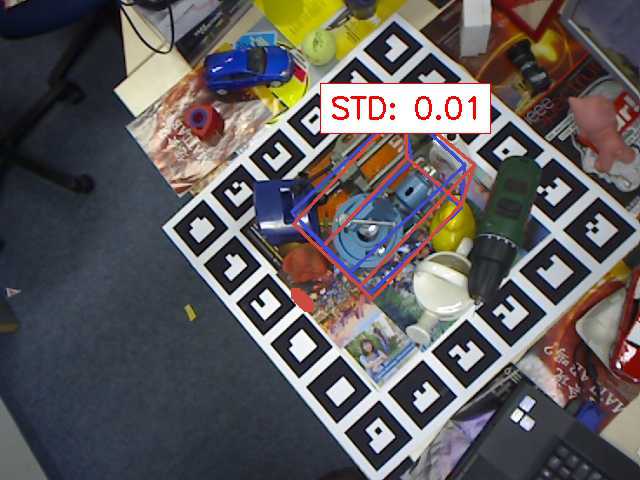}
	\includegraphics[width=.24\linewidth]{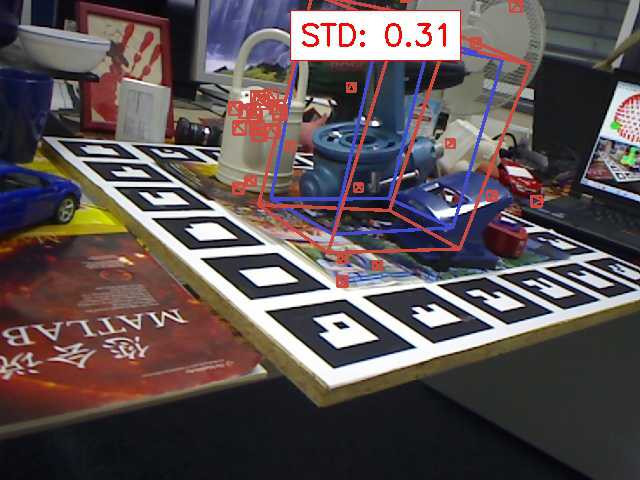} \\
	\includegraphics[width=.24\linewidth]{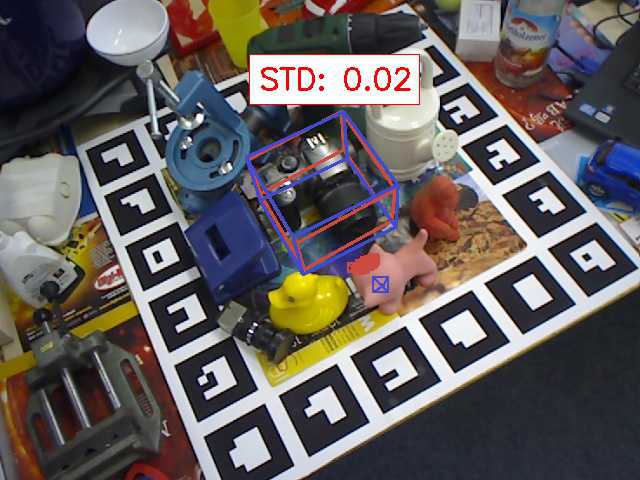}
	\includegraphics[width=.24\linewidth]{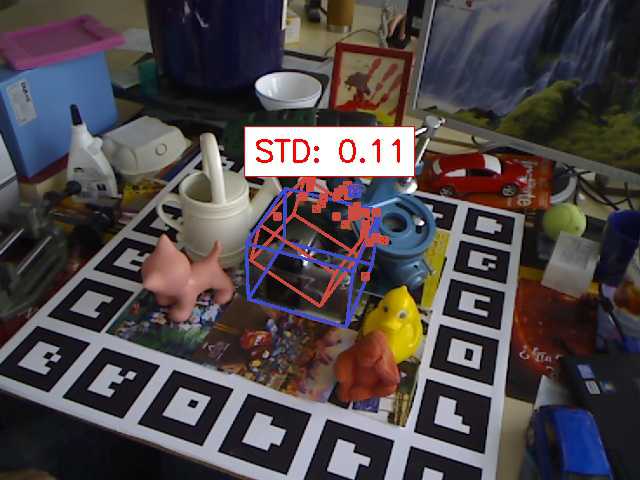}
	\hspace{3mm}
	\includegraphics[width=.24\linewidth]{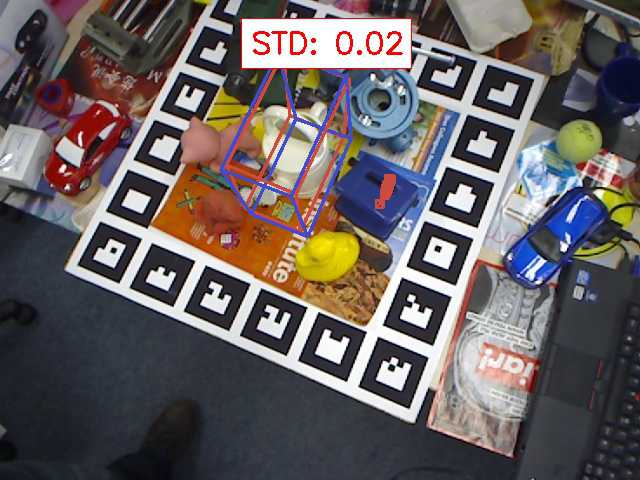}
	\includegraphics[width=.24\linewidth]{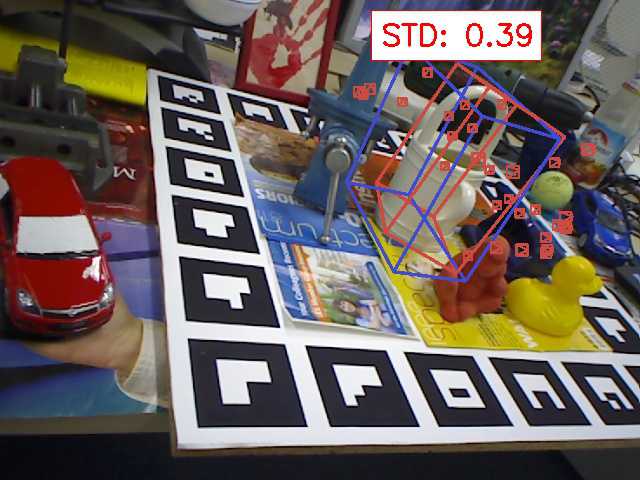} \\
	\includegraphics[width=.24\linewidth]{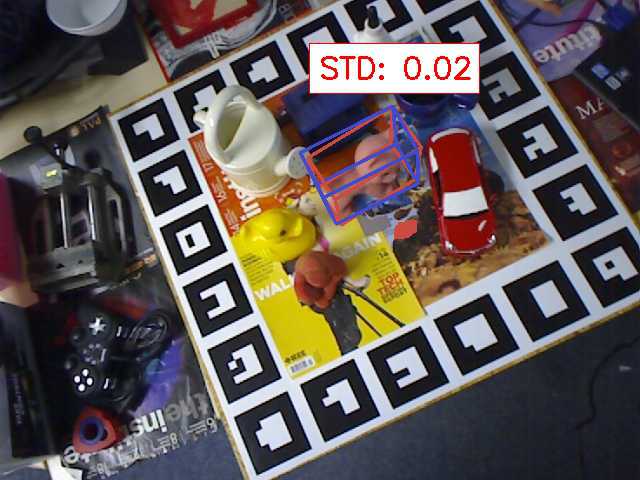}
	\includegraphics[width=.24\linewidth]{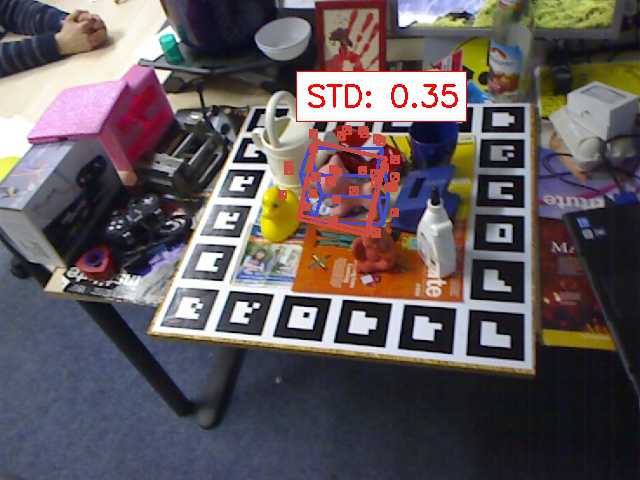}
	\hspace{3mm}
	\includegraphics[width=.24\linewidth]{figures/confidence/obj_08_lm_pose_best}
	\includegraphics[width=.24\linewidth]{figures/confidence/obj_08_lm_pose_worst}\\
	\includegraphics[width=.24\linewidth]{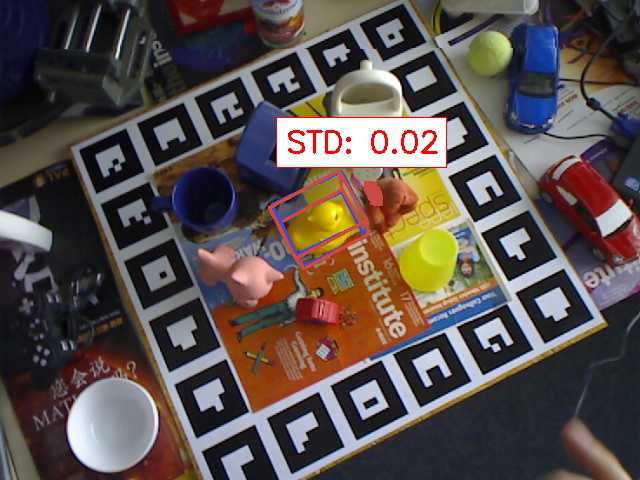}
	\includegraphics[width=.24\linewidth]{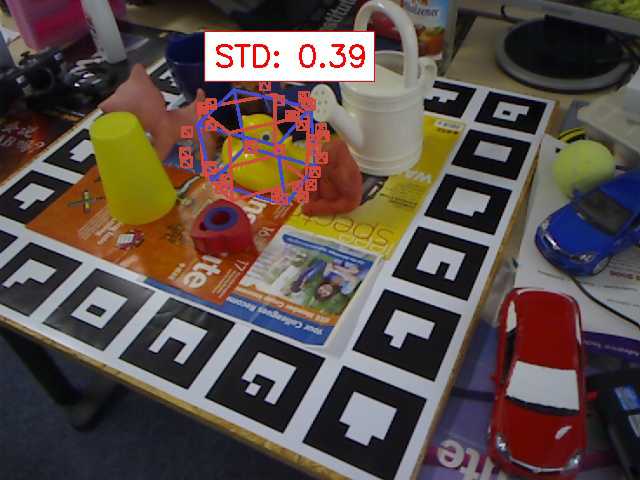}
	\hspace{3mm}
	\includegraphics[width=.24\linewidth]{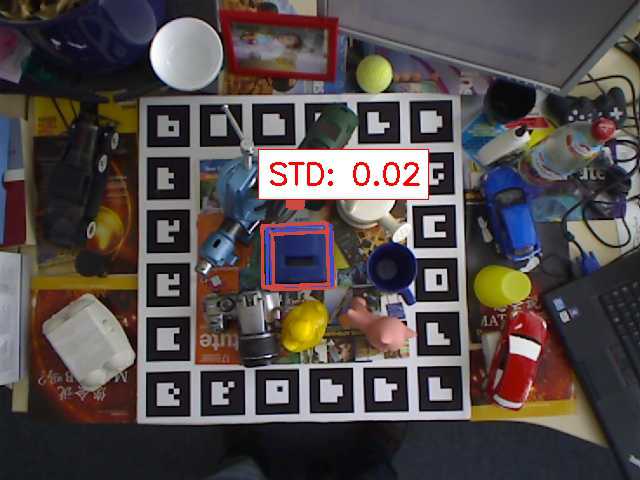}
	\includegraphics[width=.24\linewidth]{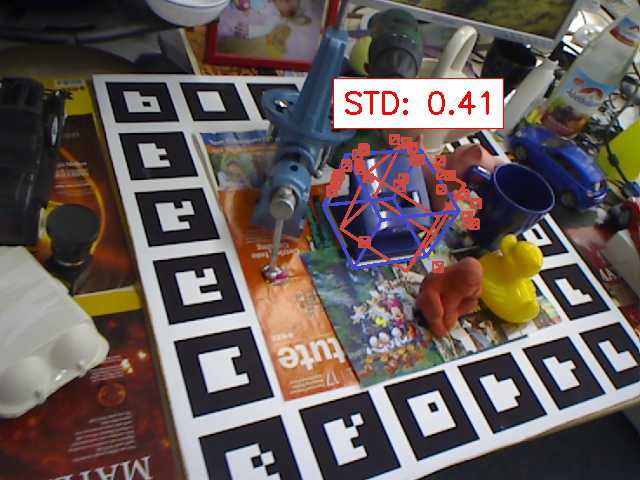}
	\centering
	\includegraphics[width=.24\linewidth]{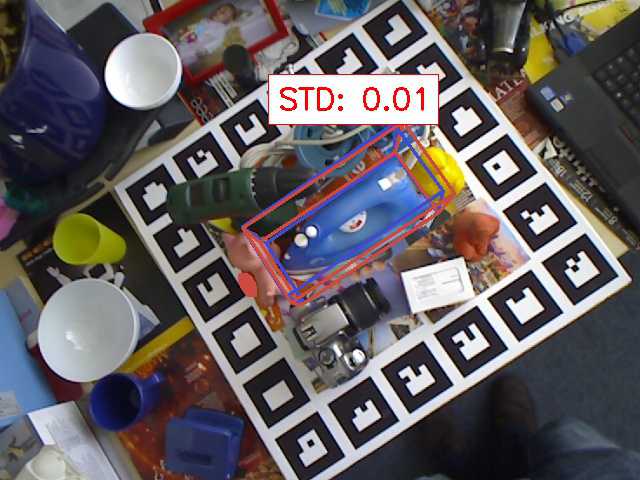}
	\includegraphics[width=.24\linewidth]{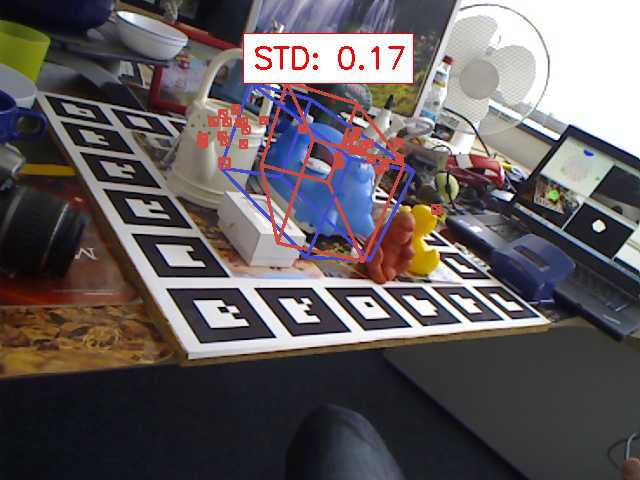}
	\hspace{3mm}
	\includegraphics[width=.24\linewidth]{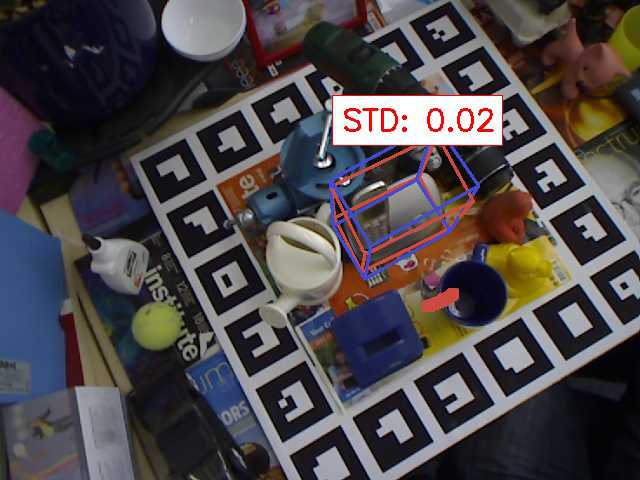}
	\includegraphics[width=.24\linewidth]{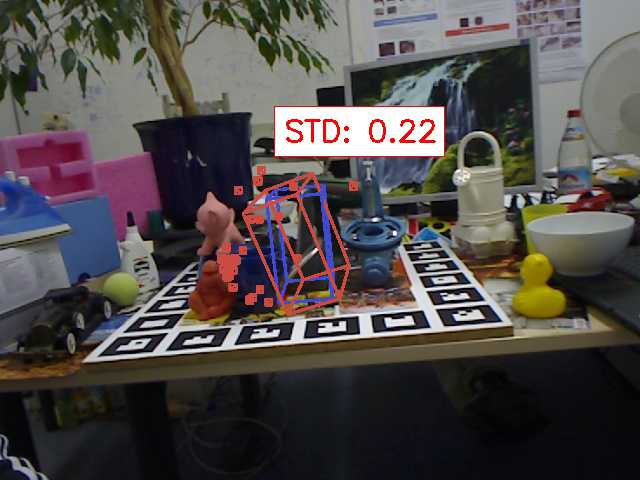}
	\caption{Qualitative examples referring to each object of the \textit{unambiguous} dataset. We show the pose with the lowest (left) and the highest (right) standard deviation in the hypotheses. Thereby, the blue bounding box depicts the ground truth pose, the red bounding box the predicted pose and the red frustums illustrate the hypotheses.}
	\label{fig:train_samples}
\end{figure}

\newpage

\section{Implementation Details}

\noindent  We implemented our method in TensorFlow \cite{Abadi2016} v1.5 and conducted all experiments on an Intel i7-5820K@3.3GHz CPU with an NVIDIA GTX 1080 GPU. We train with a batch size of 10 and use Adam \cite{Kingma2015} with a learning rate of $10^{-4}$.
\vspace{2mm}

\noindent We decay the relaxation weight $\epsilon$ from $0.05$ to $0.01$ during training (Eq. 7). Further, we empirically set $\alpha = 1.0$ and linearly increment $\beta$ from $3$ to $10$ (Eq. 8). Finally, we set $\lambda=3$, which balances rotation and translation in the final loss (Eq. 10).

\vspace{2mm}
\noindent To avoid hypotheses to \textit{die} due to bad initialization, besides sharing loss through the relaxation weight $\epsilon$, we also employ \textit{Hypotheses Dropout}: during training we deactivate a hypothesis with a probability of $p=50\%$ for the current image.

\vspace{2mm}
\noindent The mean shift and PCA implementations were taken from scikit-learn. We use \texttt{verify\_6D\_poses} in rendering/utils.py from \cite{Kehl2017}'s git repository to find the best cluster after mean shift.

\vspace{2mm}
\noindent To estimate and plot the Bingham distributions, we referred to this  \url{https://github.com/SebastianRiedel/bingham} matlab implementation. Given a set of 4D quaternions, we compute the maximum likelihood Bingham distribution employing \texttt{bingham\_fit}. We then render the sphere conducting an equatorial projection to 3D (\texttt{plot\_bingham\_3d}). Similarly, we also project the groundtruth and single hypothesis quaternions to 3D and superimpose them on the rendered sphere.

\vspace{2mm}
\noindent The pseudo-code below depicts the 6D pose inference procedure after the input image has been processed by the network.

\begin{algorithm}[h!]
\SetAlgoLined
 detections $\gets$ \{ \}\;
 \ForAll{Anchors a}{
    \vspace{2mm}
    
    \# Check If Confidence of Anchor Box is Larger Than Threshold
    
  \If{confidence(a) $\geq$ threshold}{
    \vspace{2mm}
    \# Extract Object ID and Bounding Box Center
    
   o $\gets$ object(a)\;
   $x$, $y$ $\gets$ center(a)\;
    \vspace{2mm}
    \# Extract Rotation and Depth hypotheses
    
    $\{q_j\}_{j=1}^{m}$ $\gets$ rotations(a)\;
    $\{d_j\}_{j=1}^{m}$ $\gets$ depths(a)\;
    \vspace{2mm}
    \# Principal Component Analysis on Rotation Hypotheses
    
    $\{e_j\}_{j=1}^{u}$, $\{v_j\}_{j=1}^{u}$ $\gets$ PCA($\{q_j\}_{j=1}^{m}$)\;
    
    \vspace{3mm}

   \eIf{$e_1$ $\geq$ $0.8$) }{ 
    \vspace{2mm}
        \# Pose is Ambiguous
        
        $\{c_j\}_{j=1}^{n}$ $\gets$ meanshift($\{q_j\}_{j=1}^{m}$)\;
        R, Z $\gets$ contours($\{c_j\}_{j=1}^{n}$, $\{d_j\}_{j=1}^{m}$)\;
    }{ 
    \vspace{2mm}
        \# Pose is Unambiguous
        
        R $\gets$ weiszfeld($\{q_j\}_{j=1}^{m}$)\;
        Z $\gets$ median($\{d_j\}_{j=1}^{m}$)\;
    }
    \vspace{2mm}
    \# Compute 3D Translation

    t $\gets$ backproject(camera, x, y, Z)\;
    detections.append(\{o, R, t\})\;
   }
 }
 \Return detections\;
 \caption[]{Pose Inference}
\end{algorithm}

\section{Synthetic Training Samples}

\begin{figure}[H]
	\centering
	\includegraphics[width=.24\linewidth]{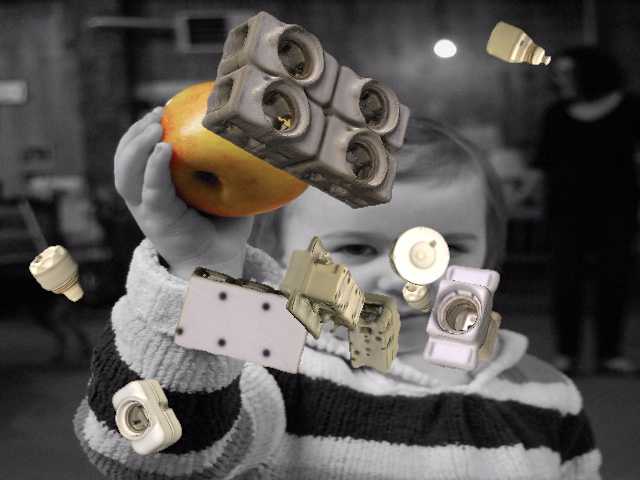}
	\includegraphics[width=.24\linewidth]{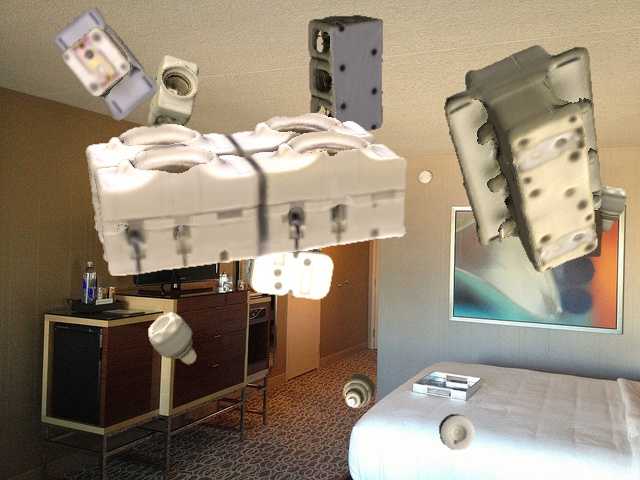}
	\includegraphics[width=.24\linewidth]{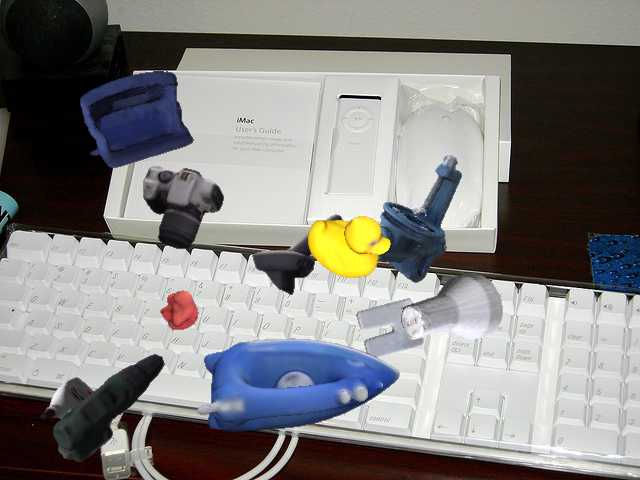}
	\includegraphics[width=.24\linewidth]{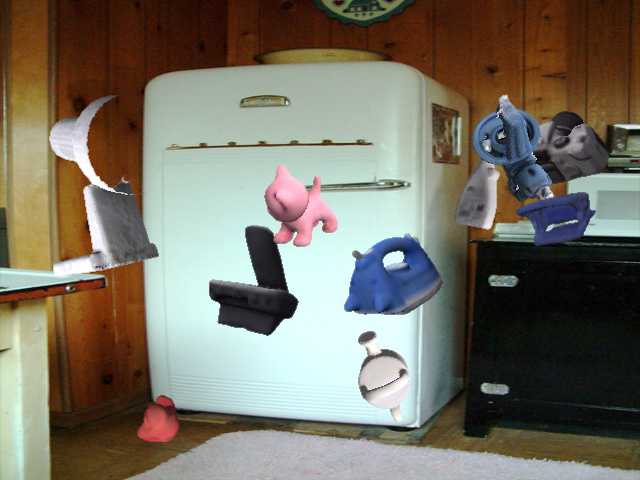}
	\caption{Example from the utilized training datasets. Left: 'T-LESS' - Right: 'LineMOD' }
	\label{fig:train_samples}
\end{figure}

We generate \textbf{synthetic samples} by rendering objects with random poses onto images from the MS COCO dataset \cite{Lin2014a}. Using OpenGL commands we generate a random pose from a valid range: 360º on the azimuth and altitude along a view sphere, and 180º for inplane rotation. We also vary the radius of the viewing sphere to enable multi-scale detection.
In order to increase the variance of the dataset, we add random perturbations such as illumination and contrast changes, among others. This is a similar approach to \cite{Kehl2017, Sundermeyer_2018_ECCV}. However, in contrast to them, for each assigned anchor box, we save exactly one 4D quaternion as the ground truth for the rotation, even if ambiguous.

\clearpage

{\small
\bibliographystyle{ieee_fullname}
\bibliography{egbib}

\begin{thebibliography}{10}\itemsep=-1pt

\bibitem{Abadi2016}
Mart{\'{i}}n Abadi, Paul Barham, Jianmin Chen, Zhifeng Chen, Andy Davis,
  Jeffrey Dean, Matthieu Devin, Sanjay Ghemawat, Geoffrey Irving, Michael
  Isard, Manjunath Kudlur, Josh Levenberg, Rajat Monga, Sherry Moore, Derek~G
  Murray, Benoit Steiner, Paul Tucker, Vijay Vasudevan, Pete Warden, Martin
  Wicke, Yuan Yu, Xiaoqiang Zheng, and Google Brain.
\newblock {TensorFlow: A System for Large-Scale Machine Learning TensorFlow: A
  system for large-scale machine learning}.
\newblock In {\em OSDI}, 2016.

\bibitem{bingham1974antipodally}
Christopher Bingham.
\newblock An antipodally symmetric distribution on the sphere.
\newblock {\em The Annals of Statistics}, pages 1201--1225, 1974.

\bibitem{birdal2018bayesian}
Tolga Birdal, Umut Simsekli, Mustafa~Onur Eken, and Slobodan Ilic.
\newblock Bayesian pose graph optimization via bingham distributions and
  tempered geodesic mcmc.
\newblock In {\em NeurIPS}, 2018.

\bibitem{Brachmann2016}
Eric Brachmann, Frank Michel, Alexander Krull, Michael Ying~Yang, Stefan
  Gumhold, et~al.
\newblock Uncertainty-driven 6d pose estimation of objects and scenes from a
  single rgb image.
\newblock In {\em CVPR}, 2016.

\bibitem{busam2017}
Benjamin Busam, Tolga Birdal, and Nassir Navab.
\newblock Camera pose filtering with local regression geodesics on the
  riemannian manifold of dual quaternions.
\newblock In {\em ICCV Workshop}, 2017.

\bibitem{Cicconet2017}
Marcelo Cicconet, Vighnesh Birodkar, Mads Lund, Michael Werman, and Davi
  Geiger.
\newblock {A convolutional approach to reflection symmetry}.
\newblock {\em PRL}, 95(1):44--50, 2017.

\bibitem{Comaniciu02meanshift}
Dorin Comaniciu, Peter Meer, and Senior Member.
\newblock Mean shift: A robust approach toward feature space analysis.
\newblock {\em TPAMI}, 24:603--619, 2002.

\bibitem{DBLP:conf/iros/CoronaKF18}
Enric Corona, Kaustav Kundu, and Sanja Fidler.
\newblock Pose estimation for objects with rotational symmetry.
\newblock In {\em IROS}, 2018.

\bibitem{Do2018}
Thanh{-}Toan Do, Ming Cai, Trung Pham, and Ian~D. Reid.
\newblock Deep-6dpose: Recovering 6d object pose from a single {RGB} image.
\newblock {\em CoRR}, abs/1802.10367, 2018.

\bibitem{Eaton2009}
Frederik Eaton and Zoubin Ghahramani.
\newblock Choosing a variable to clamp.
\newblock In {\em Artificial Intelligence and Statistics}, 2009.

\bibitem{Elawady2017}
Mohamed Elawady, Christophe Ducottet, Olivier Alata, C{\'e}cile Barat, and
  Philippe Colantoni.
\newblock Wavelet-based reflection symmetry detection via textural and color
  histograms.
\newblock {\em ICCV Workshop}, 2017.

\bibitem{glover2014tracking}
Jared Glover and Leslie~Pack Kaelbling.
\newblock Tracking the spin on a ping pong ball with the quaternion bingham
  filter.
\newblock In {\em ICRA}, 2014.

\bibitem{gramkow2001averaging}
Claus Gramkow.
\newblock On averaging rotations.
\newblock {\em Journal of Mathematical Imaging and Vision}, 15(1-2):7--16,
  2001.

\bibitem{hartley2011l1}
Richard Hartley, Khurrum Aftab, and Jochen Trumpf.
\newblock L1 rotation averaging using the weiszfeld algorithm.
\newblock In {\em CVPR}, 2011.

\bibitem{hartley2013}
Richard Hartley, Jochen Trumpf, Yuchao Dai, and Hongdong Li.
\newblock Rotation averaging.
\newblock {\em IJCVn}, 103(3):267--305, 2013.

\bibitem{He2017}
Kaiming He, Georgia Gkioxari, Piotr Doll{\'{a}}r, and Ross~B. Girshick.
\newblock Mask {R-CNN}.
\newblock In {\em ICCV}, 2017.

\bibitem{He2016}
Kaiming He, Xiangyu Zhang, Shaoqing Ren, and Jian Sun.
\newblock Deep residual learning for image recognition.
\newblock In {\em CVPR}, 2016.

\bibitem{Hinterstoisser2011}
Stefan Hinterstoisser, Stefan Holzer, Cedric Cagniart, Slobodan Ilic, Kurt
  Konolige, Nassir Navab, and Vincent Lepetit.
\newblock {Multimodal templates for real-time detection of texture-less objects
  in heavily cluttered scenes}.
\newblock In {\em ICCV}, 2011.

\bibitem{Hinterstoisser2012}
Stefan Hinterstoisser, Vincent Lepetit, Slobodan Ilic, Stefan Holzer, Gary
  Bradski, Kurt Konolige, and Nassir Navab.
\newblock Model based training, detection and pose estimation of texture-less
  3d objects in heavily cluttered scenes.
\newblock In {\em ACCV}, 2013.

\bibitem{Hinterstoisser2017}
Stefan Hinterstoisser, Vincent Lepetit, Paul Wohlhart, and Kurt Konolige.
\newblock On pre-trained image features and synthetic images for deep learning.
\newblock In {\em ECCV}, 2018.

\bibitem{Hodan2017}
Tom{\'{a}}{\v{s}} Hodan, Pavel Haluza, {\v{S}}t{\v{e}}p{\'{a}}n Obdrzalek,
  Jiř{\'{i}} Matas, Manolis Lourakis, and Xenophon Zabulis.
\newblock {T-LESS: An RGB-D dataset for 6D pose estimation of texture-less
  objects}.
\newblock In {\em WACV}, 2017.

\bibitem{Hodan2016}
Tomas Hodan, Jiri Matas, and Stepan Obdrzalek.
\newblock {On Evaluation of 6D Object Pose Estimation}.
\newblock In {\em ECCV Workshop}, 2016.

\bibitem{karcher1977}
Hermann Karcher.
\newblock Riemannian center of mass and mollifier smoothing.
\newblock {\em Communications on pure and applied mathematics}, 30(5):509--541,
  1977.

\bibitem{Ke2017}
Wei Ke, Jie Chen, Jianbin Jiao, Guoying Zhao, and Qixiang Ye.
\newblock Srn: Side-output residual network for object symmetry detection in
  the wild.
\newblock In {\em CVPR}, 2017.

\bibitem{Kehl2017}
Wadim Kehl, Fabian Manhardt, Slobodan Ilic, Federico Tombari, and Nassir Navab.
\newblock {SSD-6D: Making RGB-Based 3D Detection and 6D Pose Estimation Great
  Again}.
\newblock In {\em ICCV}, 2017.

\bibitem{Kehl2016a}
Wadim Kehl, Fausto Milletari, Federico Tombari, Slobodan Ilic, and Nassir
  Navab.
\newblock {Deep Learning of Local RGB-D Patches for 3D Object Detection and 6D
  Pose Estimation}.
\newblock In {\em ECCV}, 2016.

\bibitem{Kingma2015}
Diederik~P. Kingma and Jimmy~Lei Ba.
\newblock {Adam: a Method for Stochastic Optimization}.
\newblock In {\em ICLR}, 2015.

\bibitem{Kokkinos2016}
Iasonas Kokkinos.
\newblock Ubernet: Training a `universal' convolutional neural network for
  low-, mid-, and high-level vision using diverse datasets and limited memory.
\newblock In {\em CVPR}, 2017.

\bibitem{Krull2015}
Alexander Krull, Eric Brachmann, Frank Michel, Michael~Ying Yang, Stefan
  Gumhold, and Carsten Rother.
\newblock {Learning Analysis-by-Synthesis for 6D Pose Estimation in RGB-D
  Images}.
\newblock In {\em ICCV}, 2015.

\bibitem{kurz2013recursive}
Gerhard Kurz, Igor Gilitschenski, Simon Julier, and Uwe~D Hanebeck.
\newblock Recursive estimation of orientation based on the bingham
  distribution.
\newblock In {\em FUSION}, 2013.

\bibitem{Li2018}
Yi Li, Gu Wang, Xiangyang Ji, Yu Xiang, and Dieter Fox.
\newblock Deepim: Deep iterative matching for 6d pose estimation.
\newblock In {\em ECCV}, 2018.

\bibitem{Lin2014a}
Tsung~Yi Lin, Michael Maire, Serge Belongie, James Hays, Pietro Perona, Deva
  Ramanan, Piotr Doll{\'{a}}r, and C~Lawrence Zitnick.
\newblock {Microsoft COCO: Common objects in context}.
\newblock In {\em ECCV}, 2014.

\bibitem{Liu2016}
Wei Liu, Dragomir Anguelov, Dumitru Erhan, Christian Szegedy, Scott Reed,
  Cheng-yang Fu, and Alexander~C Berg.
\newblock {SSD : Single Shot MultiBox Detector}.
\newblock In {\em ECCV}, 2016.

\bibitem{Liu2015}
Yuanliu Liu, Zejian Yuan, Badong Chen, Jianru Xue, and Nanning Zheng.
\newblock {Illumination Robust Color Naming via Label Propagation}.
\newblock In {\em ICCV}, 2015.

\bibitem{Manhardt2019}
Fabian Manhardt, Wadim Kehl, and Adrien Gaidon.
\newblock Roi-10d: Monocular lifting of 2d detection to 6d pose and metric
  shape.
\newblock In {\em CVPR}, 2019.

\bibitem{Manhardt_2018_ECCV}
Fabian Manhardt, Wadim Kehl, Nassir Navab, and Federico Tombari.
\newblock Deep model-based 6d pose refinement in rgb.
\newblock In {\em ECCV}, 2018.

\bibitem{Ovsjanikov2008}
Maks Ovsjanikov, Jian Sun, and Leonidas Guibas.
\newblock {Global intrinsic symmetries of shapes}.
\newblock {\em Eurographics Symposium on Geometry Processing},
  27(5):1341--1348, 2008.

\bibitem{Pinto2016}
Lerrel Pinto and Abhinav Gupta.
\newblock Supersizing self-supervision: Learning to grasp from 50k tries and
  700 robot hours.
\newblock In {\em ICRA}, 2016.

\bibitem{Rad2017}
Mahdi Rad and Vincent Lepetit.
\newblock Bb8: A scalable, accurate, robust to partial occlusion method for
  predicting the 3d poses of challenging objects without using depth.
\newblock In {\em ICCV}, 2017.

\bibitem{Redmon2016}
Joseph Redmon, Santosh Divvala, Ross Girshick, and Ali Farhadi.
\newblock {You only look once: Unified, real-time object detection}.
\newblock In {\em CVPR}, 2016.

\bibitem{Ren2015}
Shaoqing Ren, Kaiming He, Ross Girshick, and Jian Sun.
\newblock Faster {R-CNN:} towards real-time object detection with region
  proposal networks.
\newblock In {\em NeurIPS}, 2015.

\bibitem{rupprecht2017learning}
Christian Rupprecht, Iro Laina, Robert DiPietro, Maximilian Baust, Federico
  Tombari, Nassir Navab, and Gregory~D Hager.
\newblock Learning in an uncertain world: Representing ambiguity through
  multiple hypotheses.
\newblock In {\em ICCV}, 2017.

\bibitem{DBLP:journals/corr/SimonyanZ14a}
Karen Simonyan and Andrew Zisserman.
\newblock {Very Deep Convolutional Networks for Large-Scale Image Recognition}.
\newblock {\em CoRR}, abs/1409.1, 2014.

\bibitem{Sundermeyer_2018_ECCV}
Martin Sundermeyer, Zoltan-Csaba Marton, Maximilian Durner, Manuel Brucker, and
  Rudolph Triebel.
\newblock Implicit 3d orientation learning for 6d object detection from rgb
  images.
\newblock In {\em ECCV}, 2018.

\bibitem{Szegedy2016}
Christian Szegedy, Sergey Ioffe, Vincent Vanhoucke, and Alex~A. Alemi.
\newblock Inception-v4, inception-resnet and the impact of residual connections
  on learning.
\newblock In {\em ICLR Workshop}, 2016.

\bibitem{43022}
Christian Szegedy, Wei Liu, Yangqing Jia, Pierre Sermanet, Scott Reed, Dragomir
  Anguelov, Dumitru Erhan, Vincent Vanhoucke, and Andrew Rabinovich.
\newblock {Going Deeper with Convolutions}.
\newblock In {\em CVPR}, 2015.

\bibitem{8100178}
Keisuke Tateno, Federico Tombari, Iro Laina, and Nassir Navab.
\newblock Cnn-slam: Real-time dense monocular slam with learned depth
  prediction.
\newblock In {\em CVPR}, 2017.

\bibitem{Tekin2018}
Bugra Tekin, Sudipta~N. Sinha, and Pascal Fua.
\newblock Real-time seamless single shot 6d object pose prediction.
\newblock In {\em CVPR}, 2018.

\bibitem{weiszfeld1937}
Endre Weiszfeld.
\newblock Sur le point pour lequel la somme des distances de n points
  donn{\'e}s est minimum.
\newblock {\em Tohoku Mathematical Journal, First Series}, 43:355--386, 1937.

\bibitem{Wohlhart2015}
Paul Wohlhart and Vincent Lepetit.
\newblock {Learning Descriptors for Object Recognition and 3D Pose Estimation}.
\newblock In {\em CVPR}, 2015.

\bibitem{Xu2018}
Bin Xu and Zhenzhong Chen.
\newblock Multi-level fusion based 3d object detection from monocular images.
\newblock In {\em CVPR}, 2018.

\bibitem{conf/cvpr/YaoFU12}
Jian Yao, Sanja Fidler, and Raquel Urtasun.
\newblock Describing the scene as a whole: Joint object detection, scene
  classification and semantic segmentation.
\newblock In {\em CVPR}, 2012.

\bibitem{Xiang2018}
Xiang Yu, Schmidt Tanner, Narayanan Venkatraman, and Fox Dieter.
\newblock Posecnn: A convolutional neural network for 6d object pose estimation
  in cluttered scenes.
\newblock In {\em RSS}, 2018.

\end{thebibliography}
}

\end{document}